\documentclass[final]{cvpr}

\usepackage{times}
\usepackage{epsfig}
\usepackage{graphicx}
\usepackage{amsmath}
\usepackage{amssymb}

\usepackage{booktabs}      
\usepackage{bm}
\usepackage{pifont}            
\usepackage{tabulary}
\usepackage[dvipsnames]{xcolor}

\usepackage{colortbl}

\definecolor{Gray1}{gray}{0.8}
\definecolor{Gray2}{gray}{0.95}

\newcommand{\cmark}{\mbox{\color{OliveGreen}\ding{52}}}%
\newcommand{\xmark}{\mbox{\color{red}\ding{56}}}%

\newcommand{\app}{\raise.17ex\hbox{$\scriptstyle\sim$}}

\newcolumntype{x}[1]{>{\centering\arraybackslash}p{#1pt}}

\newlength\savewidth\newcommand\shline{\noalign{\global\savewidth\arrayrulewidth
  \global\arrayrulewidth 1pt}\hline\noalign{\global\arrayrulewidth\savewidth}}
\newcommand{\tablestyle}[2]{\setlength{\tabcolsep}{#1}\renewcommand{\arraystretch}{#2}\centering\footnotesize}
\makeatletter\renewcommand\paragraph{\@startsection{paragraph}{4}{\z@}
  {.5em \@plus1ex \@minus.2ex}{-.5em}{\normalfont\normalsize\bfseries}}\makeatother

\usepackage[pagebackref=true,breaklinks=true,colorlinks,bookmarks=false]{hyperref}

\def\cvprPaperID{1310} 
\def\confYear{CVPR 2021}

\begin{document}

\title{Multiple Object Tracking with Correlation Learning}

\author{Qiang Wang, Yun Zheng, Pan Pan, Yinghui Xu \\
Machine Intelligence Technology Lab, Alibaba Group\\
{\tt\small \{qishi.wq, zhengyun.zy, panpan.pp, renji.xyh\}@alibaba-inc.com}
}

\maketitle

\begin{abstract}

Recent works have shown that convolutional networks have substantially improved the performance of multiple object tracking by simultaneously learning detection and appearance features.
However, due to the local perception of the convolutional network structure itself, the long-range dependencies in both the spatial and temporal cannot be obtained efficiently.
To incorporate the spatial layout, we propose to exploit the local correlation module to model the topological relationship between targets and their surrounding environment, which can enhance the discriminative power of our model in crowded scenes. 
Specifically, we establish dense correspondences of each spatial location and its context, and explicitly constrain the correlation volumes through self-supervised learning.
To exploit the temporal context, existing approaches generally utilize two or more adjacent frames to construct an enhanced feature representation, but the dynamic motion scene is inherently difficult to depict via CNNs. Instead, our paper proposes a learnable correlation operator to establish frame-to-frame matches over convolutional feature maps in the different layers to align and propagate temporal context.
With extensive experimental results on the MOT datasets, our approach demonstrates the effectiveness of correlation learning with the superior performance and obtains state-of-the-art  MOTA of $76.5\%$ and IDF1 of $73.6\%$ on MOT17.

\end{abstract}

\section{Introduction}
\label{sec:intro}

Multi-Object Tracking (MOT) is an essential component for computer vision with many applications, such as video surveillance~\cite{surveillance} and modern autonomous driving~\cite{autodrive,waymo}.
It aims to continuously locate trajectories of multiple targets in video frames.
Decades of research efforts have led to impressive performance on challenging benchmarks~\cite{mot15,mot16,mot20}.

MOT has traditionally adopted the tracking-by-detection paradigm~\cite{sort,mpn,tracktor,fairmot}, which capitalizes on the natural division of detection and data association tasks for the problem.
These algorithms extract appearance features within each detection patches and record object location information for subsequent data association~\cite{deepsort,mpn}.
This tracking paradigm makes researchers mainly focus on optimizing detection~\cite{poi,bjd}, feature representation~\cite{motdt,tmlf}, or data association~\cite{sort,mpn,lift}.
With the rapid progress of detection algorithms~\cite{dpm,fastrcnn,fasterrcnn,yolov3}, the detection based tracking has achieved great performance improvement~\cite{poi,bjd}.
Although tremendous strides have been made in MOT, there still exists tough challenges in determining distractors and frequent occlusions, especially in complex interactive scenes~\cite{mot20}.
Additionally, the above cascaded structure is inefficient and prevents the joint optimization between stages.

\begin{figure}[t]
\includegraphics[width=0.5\textwidth]{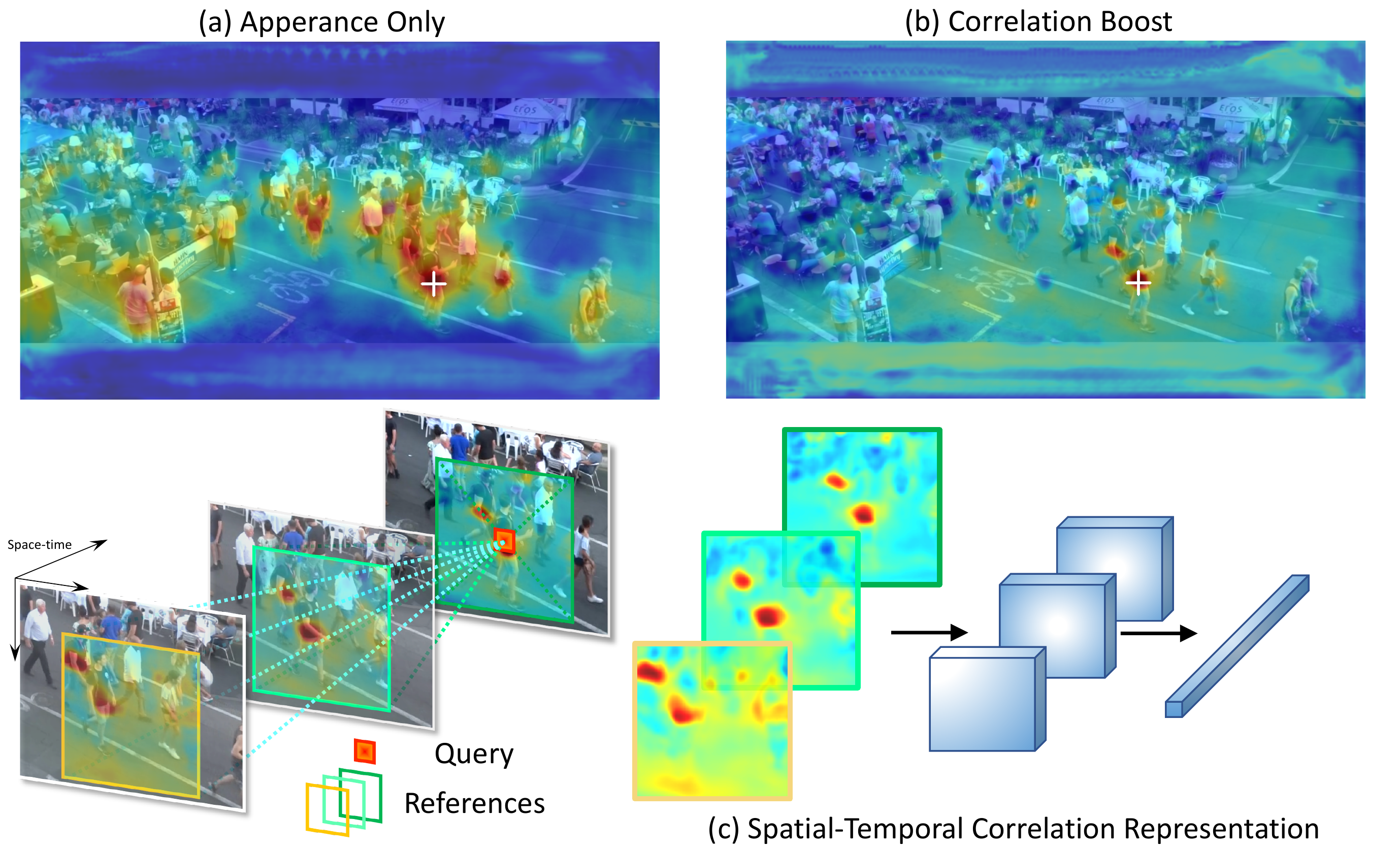}
\caption{Visualization of the matching confidences (a)-(b) computed between the indicated target (white cross) in the reference image and all locations of the query image. The appearance feature based tracker~\cite{fairmot} (a) generates undistinctive and inaccurate confidences due to the existing of similar distractors. 
In contrast, our Correlation Tracker (b) predicts a distinct high-confidence value at the correct location, with correlation learning (c).}
\vspace{-3mm}
\label{fig:order}
\end{figure}

One promising approach is to extend the end-to-end trainable detection framework~\cite{fasterrcnn,yolov3,centertrack} to jointly learn detection and appearance feature, which has largely advanced the state-of-the-art in MOT~\cite{mots,jde,retinatrack,fairmot}.
However as illustrated in Fig.~\ref{fig:order}, in the case of existing similar distractors, the appearance feature generates undistinctive  and inaccurate matching confidences (Fig.~\ref{fig:order}a), severely affecting the performance of association. 
These methods are limited in local descriptors, and it is difficult to distinguish similar objects. While as shown in Fig.~\ref{fig:order}c, the context relation map can help to easily distinguish different targets.

Based on those observations, we propose a correlation network to learn the topological information of the object and context.
Specifically, we use a spatial correlation layer to record the relationship between targets and relative spatial positions. 
While constructing a full correlation (\eg,~non-local~\cite{nonlocal}) for all locations is computationally prohibitive for real-time MOT, this work constructs a local correlation volume by limiting the search range at each feature pyramid.
Besides, our correlation learning is not limited for targets of interest category~\cite{strmot,jdmotgnn}. 
Background contexts, such as vehicles, are also modeled to help target recognition and relational reasoning (Fig.~\ref{fig:order}c).
We establish dense correspondences of each spatial location and its context, and explicitly constrain the correlation volumes through self-supervised learning.

Further, the detector in MOT usually uses independent frames as input and therefore does not make full use of temporal information. This detection method makes the algorithm suffer from missing detection in crowded scenes, and further increases the difficulty of subsequent data association.
Recently, adjacent frames~\cite{centertrack, ctracker} or three frames \cite{tubetk} are adopted to enhance the temporal consistence. 
The performance of the algorithm in occlusion scenes has been improved to a certain extent, but these methods are still limited with fewer frames. 
CenterTrack~\cite{centertrack} attempt to use an aggressive data augmentation to increase the ability of target alignment, but convolution networks itself are inherently limited in local receptive fields. 
To solve the above problem, we extend the spatial correlation module to the temporal dimension and incorporate the historical information to reduce ambiguities in object detections.

To summarize, we make the following contributions:
\begin{itemize}
\setlength\itemsep{-0.5em}
\item We propose CorrTracker, a unified correlation tracker to intensively model associations between objects and transmit information through associations.
\item We propose a local structure-aware network and enhance the discriminability of similar objects with self-supervised learning.
\item We extend the local correlation network to model temporal information efficiently.
\item CorrTracker shows significant improvements over existing state-of-the-art results in four MOT benchmarks. In particular, we achieve $76.5\%$ MOTA and IDF1 of $73.6\%$ on MOT17.

\end{itemize}

\section{Related Work}
\label{sec:related}

\noindent \textbf{Real-time Tracking.}
As MOT has strong practical merit, the tracking speed attracts much attention. 
The researchers start from the simplest IOUTracker~\cite{ioutracker}, which only uses the intersection-over-union of bounding boxes for tracking, to add the motion model of Kalman Filter~\cite{sort} to predict the position of the rectangular boxes for matching.
Although they have achieved amazing speed, stable tracking cannot be achieved under challenges such as target interleaving. 
Researchers~\cite{motdt,deepsort} introduce Person Re-Identification (ReID) features as an appearance model to increase the discriminative power of the tracker.
However, the individual calculation for patches makes the object classification and ReID feature extraction as a computational bottleneck. 
MOTDT~\cite{motdt} achieves real-time tracking by using RoI-pooling~\cite{fastrcnn} on a shared feature map. 
In order to further decrease the computational cost of ReID feature extraction, JDE~\cite{jde} adds a ReID branch in a single-stage detector YOLOv3~\cite{yolov3} to achieve efficient ReID feature calculation. 
FairMOT~\cite{fairmot} explores the importance of detection and recognition tasks and uses anchor-free method~\cite{centernet} to reduce the ambiguity of anchors.
We are mainly based on FairMOT, which achieves the state-of-the-art performance with a more balanced ReID and detection.

Other researchers~\cite{tracktor, centertrack,ctracker,tubetk} explore new tracking paradigms to remove ReID recognition. 
Tracktor~\cite{tracktor} uses the the bounding boxes in the previous frame to directly regress the current locations.
CenterTrack~\cite{centertrack}, ChainedTracker~\cite{ctracker}, and TubeTK~\cite{tubetk} use multiple frames to simultaneously predict the bounding boxes for adjacent frames to achieve short association, thereby merging to long-term tracks.
However, these methods usually have many identity switches because they cannot model long-term dependencies.


\begin{figure*}[t]
\includegraphics[width=0.99\textwidth]{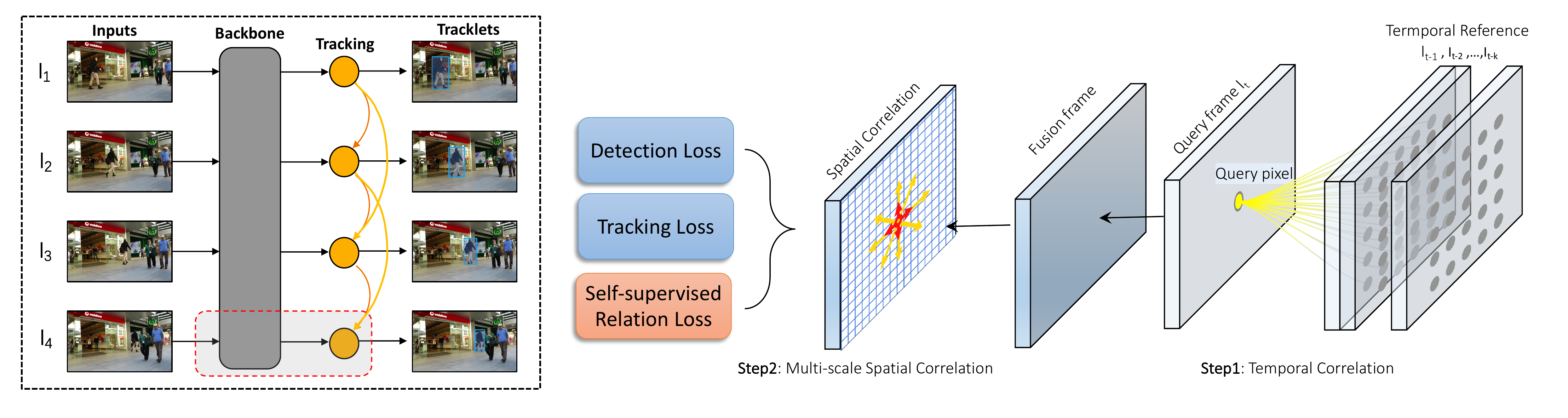}
\caption{ \textbf{Overview}. We enhance the appearance features with correlation layer, which densely encoding pairwise relation of object and their spatial-temporal neighbourhoods. The local correlation volumes are optimized in a self-supervised manner.}
\label{fig:pipeline}
\end{figure*}

\noindent \textbf{Tracking with Graph Model.}
MOT has traditionally been approached as a hand-crafted graph optimization problem~\cite{costflow, lift}, where the cropped targets are treated as nodes.
Recently, graph neural network based methods~\cite{strmot,jdmotgnn,mpn} have been shown as a promising alternative to traditional optimization methods.
State-of-the-art approach~\cite{strmot} utilizes graph convolutional network to propagate features in the spatial-temporal space.
MPN~\cite{mpn} introduces the message passing network to dissect the information and associate detections through the edge classification.
Different from these methods, feature propagation is carried out at the frame feature level, which can absorb the information of both the foreground and background and reduce the loss of contextual information.

\noindent \textbf{Tracking with Optical Flow.}
FlowTrack~\cite{flowfuse} introduces optical flow to predict the target location.
But explicitly using optical flow is not only time-consuming, but also only encodes the pixel-level motion.
CenterTrack~\cite{centertrack} borrows the method of optical flow to directly predict the movement of the target center between two frames, which is called \textit{instance flow}.
However, directly predicting the offset on the concatenated feature map needs to provide training samples with all displacement, which requires excessive data augmentation.
Our correlation method predicts a dense set of matching confidence for each target, which is intrinsically invariant to translation of the paired frames. 
Our correlation operation is similar to the correlation volume in optical flow~\cite{flownet,pwc,raft} and correlation filter~\cite{famnet}. 
We both predict dense local correlation, and regard it as a part of the feature description. 
However, optical flow does not calculate the internal correlation of the image, nor does it have propagated the feature from multiple frames.
D{\&}T~\cite{dt} also utilizes correlation layer to predict candidate motion between pair of consecutive frames. Compared with it, our anchor-free framework is more compact and efficient.

\noindent \textbf{Attention Mechanism.}
Our modeling of local correlation is similar to the self-attention mechanism and Transformer.
Transformer has been a huge success in the NLP field~\cite{attention} and has also been adapted to the computer vision~\cite{nonlocal,relationnet,cosnet} to capture long-range dependencies. 
In order to reduce the quadratic complexity of the non-local operation, the researchers propose to shrink the attention span with local region~\cite{imgtf}, or only along individual axes~\cite{axial}. 
Different from these methods, we mainly encode the context identity through local correlation weighting, and use this cues to increase the model robustness.

\section{Methodology}
\label{sec:method}

Figure~\ref{fig:pipeline} shows the overall pipeline of the proposed CorrTracker. 
Our method can be distilled down to three stages: (1) general feature extraction, (2) simultaneous learning correlation from spatial-temporal dependencies and predicting the detection, and (3) performing data association to assign detections into their most likely trajectories, where stage (1) and stage (2) are differentiable and composed into an end-to-end trainable architecture. 
We adopt a compact association technique that is similar to the one used by DeepSORT~\cite{deepsort} to control the initialization and the termination of tracks. 
The main contribution is the highly efficient modeling for the correlation between dense location and their context on feature maps, which helps suppressing distractors in complex scenes.

\subsection{Motivation}
For each input video frame $\mathbf{I}_t \in \mathbb{R}^{H\times W \times 3}$, an object detector is applied to find all candidate detections $\mathcal{D}_t = \{ \mathbf{d}_t^i \}_{i=1}^{N}, \mathbf{d}_t^i = ( x_t^i,  y_t^i,  w_t^i,  h_t^i)$ appearing in this frame and we have existing trajectories $\mathcal{T} _{t-1}= \{ \mathbf{T}_{t-1}^j \}_{j=1}^{M}, \mathbf{T}_{t-1}^j =\{\mathbf{d}_{1}^j,..., \mathbf{d}_{t-2}^j,\mathbf{d}_{t-1}^j\}$.
Then the affinity matrix $\mathbf{A} \in \mathbb{R}^{N \times M}$ is estimated by pair-wise comparisons of  cropped patches and existing trajectories.
The metric jointly considers both the appearance features $\mathbf{f}(\cdot) \in \mathbb{R}^{d}$ and geometric representations.
\begin{equation}
{\mathbf{A}}_{ij} ={\rm dist}(\mathbf{f}(\mathbf{d}_t^i), \hat{\mathbf{f}}(\mathbf{T}_{t-1}^j)) + \alpha {\rm IoU}(\mathbf{d}_t^i, \hat{\mathbf{d}}_t^j),
\end{equation}
The discriminative feature $\hat{\mathbf{f}}(\mathbf{T}_{t-1}^j)$ of a trajectory is usually updated with a constant-weighting strategy to follow the appearance changes.
Each confidence value for appearance feature is obtained in a distance metric, \emph{e.g.}, the inner product space.
However, the sole reliance on person-to-person feature comparisons are often insufficient to disambiguate multiple similar regions in an image. 
As illustrated in Fig.~\ref{fig:order}, in the case of similar distractors, the feature extractor usually generates inaccurate and uninformative matching confidences (Fig.~\ref{fig:order}a), severely affecting the performance of data association. 
This is the key limitation of appearance feature matching, since co-occurring similar objects are all pervasive in MOT.

Patch based feature extraction is applied as a prevalent scheme in MOT owing to its intuition. 
However the correlation information between the cropped image patches is lost directly, and the adjacency spatial relationship is only retained in coordinates $\mathbf{d}_t^i$. 
Although the subsequent data association will be globally optimized, directly using ReID features without considering the context tends to introduce more identity switches, lagging the tracking robustness.
To deal with this problem, we model the local structure of objects to distinguish it from distractors.

Inspired by correlation volume from optical flow~\cite{flownet}, we observe that a confidence value in the correlation volume models the geometric structure of each target. 
We design a novel dense correlation module, aiming to explore the context information for MOT. 
The relative position is encoded in the correlation volumes, which can be used as an auxiliary discriminant information.

\subsection{Spatial Local Correlation Layers}
\label{sec:slcl}
In this work, we use Spatial Local Correlation Layers to model the relational structure for associating a target with its neighbour.
In our local correlation layer, the feature similarity is only evaluated in the neighbourhood of the target image coordinate. 
Formally, we let $l$ denote the level in the feature pyramid and the correlation volume $\mathbf{C}^l$ between the query feature $\mathbf{F}_q^l \in \mathbb{R}^{H_l\times W_l \times d_l}$ and reference feature $\mathbf{F}_r^l \in \mathbb{R}^{H_l\times W_l \times d_l}$ is defined as,
\begin{equation}
\label{eq:dist}
\mathbf{C}^l(\mathbf{F}_q, \mathbf{F}_r, \mathbf{x}, \mathbf{d}) = \mathbf{F}_q^l (\mathbf{x})^{T} \mathbf{F}_r^l (\mathbf{x}+\mathbf{d}),  \Vert\mathbf{d}\Vert_{\infty} \leq R, 
\end{equation}
where $\mathbf{x} \in \mathbb{Z}^2$ is a coordinate in the query feature map and $\mathbf{d} \in \mathbb{Z}^2$ is the displacement from this location. 
The displacement is constrained to $\Vert\mathbf{d}\Vert_{\infty} \leq R$ , \emph{i.e.} the maximum motion in any direction is $R$. 
While most naturally thought of as a 4-dimensional tensor, the two displacement dimensions are usually vectorized into one to simplify further processing in the CNN. 
The resulting 3-d correlation volume $\mathbf{C}^l$ thus is of size $H^l \times W^l \times (2R + 1)^2$.
We also introduce the dilation tricks~\cite{dilated}, which can increase the receptive field without additional cost.
We use element-wise addition to incorporate the correlation feature into a unified appearance representation.
This context correlation features are encoded by a feed-forward Multilayer Perceptron (MLP) to match the number of channels $d_l$ in appearance features $\mathbf{F}_t^l$.
\begin{equation}
\mathbf{F}_{\mathbf{C}}^l = \mathbf{F}_t^l + {\rm{MLP}}^l\left(\mathbf{C}^l\right(\mathbf{F}_t^l, \mathbf{F}_t^l)).
\end{equation}
The non-local~\cite{nonlocal} module is to explicitly model all pairwise interactions between elements in a feature maps $\mathbf{F}_t^l \in \mathbb{R}^{H^l\times W^l \times d^l} $. 
The resulting four-dimensional correlation volume $NL(\mathbf{F}_t^l) \in \mathbb{R}^{H^l\times W^l \times H^l\times W^l} $ captures dense matching confidences between every pair of image locations. 
They build a full connection volume at a single scale, which is both computationally expensive and memory intensive. 
By contrast, our work shows that constructing a local correlation volume leads to both effective and efficient models.
In comparison with the global correlation method, our local correlation model adds less overhead to the latency (see Table~\ref{tab:archablation}).

\begin{figure}
\includegraphics[width=0.5\textwidth]{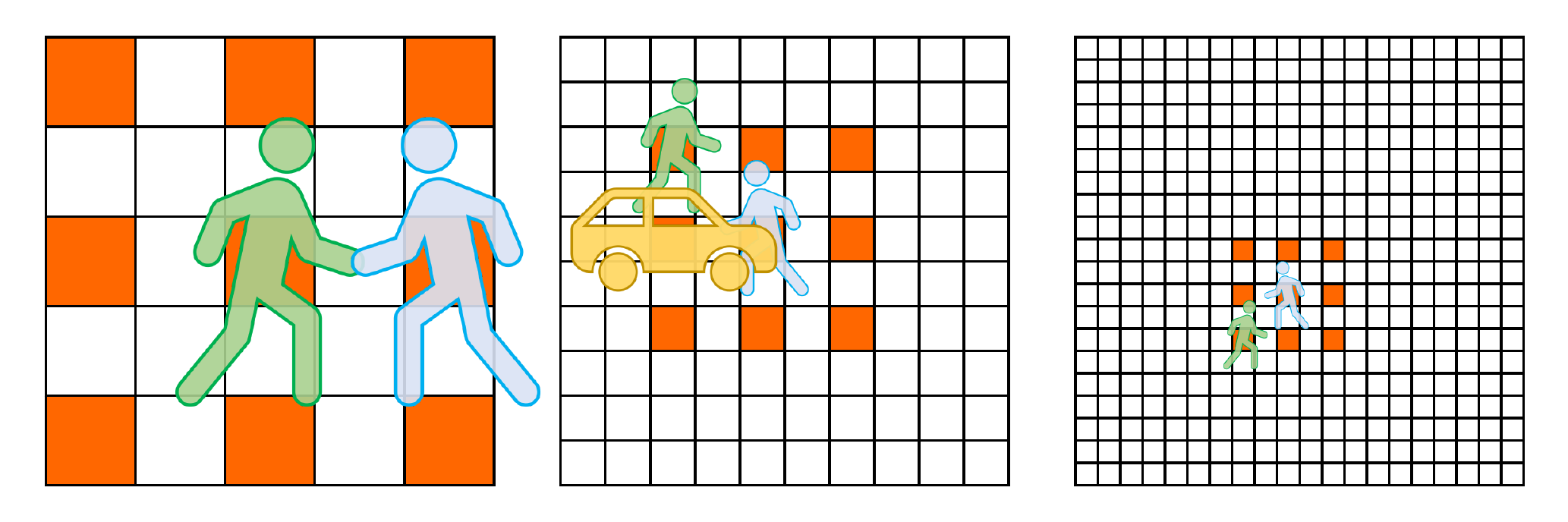}
\caption{Correlation at Multiple Pyramid Levels. For a feature tensor in $\mathbf{F}^l$, we take the inner product with local region ($R=1, D=2$) to generate a 3-d correlation volume $W^l\times H^l\times(2R+1)^2$, $D$ is the dilation rate.}
\label{fig:relation}
\end{figure}

\subsection{Correlation at Multiple Pyramid Levels}
In order to achieve long-range correlation, we propose to learn correlation at feature pyramids, as shown in Figure~\ref{fig:relation}.
On the one hand, we hope that our correlation module can obtain long-distance dependencies as much as possible, but as the local region size $R$ increases, both calculation and storage increase significantly, which hinders the application.
On the other hand, MOT naturally needs to deal with multi-scale targets.
The two-stage detection~\cite{fasterrcnn} uses RoI pooling~\cite{fastrcnn} to eliminate the difference in target scales, but this type of method usually suffers from high processing latency.
In order to solve the above problems, we utilize the general pyramid structure in the convolutional network  and learn correlation on the feature pyramids. 
Our multi-scale pyramid correlation can also be regarded as a comparison of multi-granularity features, covering the spatial context in the range of $[0, R\times D\times 2^l]$, where $D$ refers to the dilation rate. 
And, we pass this correlation from the top layer to the bottom layer,
\begin{equation}
 \hat{\mathbf{F}}_{\mathbf{C}}^{l-1} = \mathbf{Conv}(\mathbf{Upsample}(\mathbf{F}_{\mathbf{C}}^l))+\mathbf{F}_{\mathbf{C}}^{l-1},
\end{equation}
In this way, we can obtain an approximate correlation between the target and the entire global context, while keep the compactness and efficiency.
Our pyramid correlation leverages the natural spatial-temporal coherence in videos.
Multi-object tracking can be decomposed into multiple independent single-object tracking. 
Our method can be equivalent to a dense siamese network tracking~\cite{siamfc} on the feature pyramid.
On the other hand, from the perspective of set matching, global characteristics need to be considered. 
Our multi-scale correlation takes into account both aspects of information transmission.

\subsection{Temporal Correlation Learning}
\label{sec:tlcl}
The correlation between different frames are usually ignored by the MOT field, and trackers usually overcome occlusion through data association. 
Single frame detector is difficult to ensure a good temporal consistency~\cite{stability}.
This makes the algorithm's performance drop significantly in occlusion, motion blur and small target scenes, which becomes the bottleneck for MOT.
We extend the spatial local correlation from Section~\ref{sec:slcl} to the temporal dimension, and establish correlation for the targets in different frames. 
The correlation between two frames can be viewed as the establishment of motion information learning. 
We also use this correlation to enhance the feature representation, which can increase the detection accuracy.

Specifically, we establish multi-scale correlation between different frames, and use reference images as memory to enhance image features. 
This method helps tracker overcome target occlusion and motion blur, and increases the consistency of detection and identity features.
\begin{eqnarray}
\hat{\mathbf{F}}_q (\mathbf{x}) = \sum_{\forall \left \| \mathbf{d} \right \|_{\infty} < R}^{}  \frac{\mathbf{C}^l(\mathbf{F}_q, \mathbf{F}_{r},\mathbf{x},\mathbf{d})}{ (2R+1)^2 } \mathbf{F}_r(\mathbf{x}+\mathbf{d})
\\
\mathbf{C}^l(\mathbf{F}_q, \mathbf{F}_{r},\mathbf{x},\mathbf{d}) = \mathbf{F}_q^l (\mathbf{x})^{T}\mathbf{F}_{r}^l (\mathbf{x}+\mathbf{d}),  \Vert\mathbf{d}\Vert_{\infty} \leq R
\end{eqnarray}

Similar to the multi-head attention~\cite{attention}, we adopt the embedded features and dot-product similarity. 
In our case, we set the normalization factor as $(2R+1)^2$ and locally aggregate features.
This shrinked region design also comes from the motion prior of the MOT scene.
For the minimal memory consumption and fastest run-time, we can only save the previous features $\mathbf{F}_{t-1}$ in the memory. 
For the maximum accuracy, our long-term model saves the latest 5 frame image features by default.

\subsection{Self-supervised Feature Learning}
\label{sec:self}
In Section~\ref{sec:slcl} and Section~\ref{sec:tlcl}, we present how we model the correlation in spatial and temporal dimension. 
We can simply use the proposed correlation module as a plugin module without explicitly adding constraints, similar to the non-local module, which has shown significant improvement. 
Here we investigate a multi-task learning approach that imposes a semantic supervision from visual object tracking~\cite{siamfc} and self-supervised training from correspondence flow~\cite{trackcolor} on correlation volumes.

Our correlation module is interpretable, measuring the similarity between different objects. 
Actually, our method intensively performs $M\times N$ siamese tracking operations~\cite{siamfc} to increases the discrimination.
In this view, we can explicitly impose a tracking supervision. Specifically, we set up the ground-truth label as
\begin{equation}
\tilde{C}^l(\mathbf{F}_q, \mathbf{F}_{r},\mathbf{x},\mathbf{d})=\left\{\begin{matrix}
1~\mathbf{if} ~\mathbf{y}_{q}(\mathbf{x}) = \mathbf{y}_{r} (\mathbf{x}+\mathbf{d}) \\ 
0~\mathbf{if} ~\mathbf{y}_{q}(\mathbf{x}) != \mathbf{y}_{r} (\mathbf{x}+\mathbf{d}) ,\\ 
-1~\mathbf{if} ~\mathbf{y}_{q}(\mathbf{x}) < 0
\end{matrix}\right.
\end{equation}
where $\mathbf{y}$ is the identity label of the corresponding position in feature maps. 
We ignore the position without objects $\mathbf{y}_{q}(\mathbf{x}) < 0$ and use a class-balanced cross-entropy loss~\cite{siamfc}.

Inspired by the recent advances of self-supervised tracking~\cite{trackcolor}, we use \textit{colorization} as a proxy task for training our local correlation.
\begin{equation}
\hat{\mathbf{I}}_{q}(\mathbf{x}) = \sum_{\forall \left \| \mathbf{d} \right \|_{\infty} < R}^{}  \frac{\mathbf{C}^l(\mathbf{F}_q, \mathbf{F}_{r},\mathbf{x},\mathbf{d})}{(2R+1)^2}\mathbf{I}_r(\mathbf{x}+\mathbf{d}),
\end{equation}
we use the cross-entropy categorical loss after quantizing the color-space into discrete categories~\cite{trackcolor}.

\subsection{Tracking Framework}
We modify the FairMOT~\cite{fairmot} backbone by adding correlation module before the iterative deep aggregation module~\cite{dla}.
Our model retains the detection and ReID branches, and adding correlation loss in Sec.~\ref{sec:self} for multi-task learning.
For the tracking inference, our tracker first calculates the similarity between the detections of the current frame and the previous trajectories according to Eq.~(\ref{eq:dist}), and use the Hungarian algorithm~\cite{hungarian} for finding the optimal matching. 
The unmatched detections are used to initialize new trajectories. 
In order to reduce false positives, we mark these new trajectories as ``\textit{inactive}'' until the next frame is matched again and confirmed as ``\textit{active}''.
The unmatched trajectories are set to the ``lost'' state. 
When the continuous lost time $t_{loss}$ of a trajectory exceeds the threshold $\tau_{loss}$, we put it in the remove set.
If there is a success matching before removing, we restore the trajectory to the active state.
We use Kalman Filter~\cite{kalman} to model the pedestrian motion and keep the same settings in FairMOT~\cite{fairmot}.

\section{Experiments}
\label{sec:experiments}
To demonstrate the advantages of the proposed correlation tracker, we first compare the correlation module with other relational reasoning methods~\cite{nonlocal,corrnet} and evaluate different settings to justify our design choices in Section~\ref{sec:ablation}. 
Then we show that our correlation tracker outperforms the state-of-the-art methods on four MOT benchmarks~\cite{mot15,mot16,mot20} in Section~\ref{sec:sota}. 
Finally, we visualize the tracking trajectories in Section~\ref{sec:vis} and compare with other motion prediction based trackers~\cite{centertrack,ctracker,tubetk}.

\subsection{Implementation Details}
\textbf{Network Setup.}
The implementation and hyper-parameters mostly follow~\cite{fairmot}, we adopt CenterNet~\cite{centernet} detector with a variant of Deep Layer Aggregation (DLA-34)~\cite{dla} as backbone and utilize the iterative deep aggregation module (IDA) to recover a high-resolution feature map with stride $4$.
We also add a $3\times 3$ deformable convolution layer~\cite{dcn} before every upsampling stage.
The backbone network is initialized with the parameters pre-trained on COCO~\cite{coco} and then pre-trained on CrowdHuman~\cite{crowdhuman} with self-supervised learning as FairMOT~\cite{fairmot}.
The proposed correlation module is augmented before IDA module to fuse multi-scale correlation. 
For the correlation module, we set local region size $R = 5$ and dilation rate $D = 2$.

\textbf{Training and Validation Datasets.}
For a fair comparison, we also use the default training datasets as FairMOT~\cite{fairmot}.
There are six training datasets including the ETH~\cite{eth}, CityPerson~\cite{cityperson}, CalTech~\cite{caltech}, CUHK-SYSU~\cite{cuhksysu}, PRW~\cite{prw} and MOT17~\cite{mot16}.
ETH and CityPerson only provide box annotations, so we ignore the ReID losses from these datasets.
CalTech,  CUHK-SYSU, PRW and MOT17 provide both box and identity annotations which allows us to train both branches. 
We de-duplicate the overlaps between ETH datasets and MOT16 for fair comparison. 
For all validation experiments, we use the six datasets mentioned above and the first half frames of MOT17 as training, and the second half of MOT17 as the validation set.

We train on an input resolution of $1088 \times 608$, which yields an output resolution of $272 \times 152$. 
We use random flip, random scaling (between 0.5 to 1.2), cropping, and color jittering as data augmentation, and use Adam~\cite{adam} to optimize the overall objective. 
The learning rate is initialized as $1e^{-4}$ and then decayed to $1e^{-5}$ in the last 10 epochs.
We train with a batch-size of 12 (on 2 GPUs) for 30 epochs.
In the training phase, we sample 5 temporally ordered frames within a random interval of less than 3.

\textbf{Test Datasets and Evaluation Metrics.}
We evaluate the performance of our correlation tracker on the 2DMOT2015~\cite{mot15}, MOT16~\cite{mot16}, MOT17~\cite{mot16}, and MOT20~\cite{mot20}.
In particular, 2DMOT2015~\cite{mot15} contains 11 test videos.  MOT16~\cite{mot16} and MOT17~\cite{mot16} contain same 7 test videos with part different annotations.
The MOT20~\cite{mot20} consists of 4 test videos on extremely crowded scenes, which makes them really challenging.

We adopt the standard metrics of MOT Benchmarks for evaluation, including Multiple Object Tracking Accuracy (MOTA)~\cite{clear}, 
ID F1 Score~\cite{idf1}, 
Mostly tracked targets (MT), 
Mostly lost targets (ML), 
the number of False Positives (FP), 
the number of False Negatives (FN), 
and the number of Identity Switch (ID Sw.)~\cite{ids}. 
The run time is also provided and evaluated on a NVIDIA Tesla V100 GPU.

\subsection{Ablation Studies} 
\label{sec:ablation}
To elaborate on the effectiveness of the proposed approach, we conduct extensive ablation studies. 
First, we give detailed correlation analysis with different settings to justify our design choices, as presented in Table.~\ref{tab:archablation}. 
Next, the tracking accuracy and runtime for different region sizes of the correlation module is explored. 
Different building blocks are compared to illustrate the effectiveness and efficiency of the full correlation tracker.

\begin{table}[t]
\caption{Evaluation of correlation architecture on the MOT17~\cite{mot16} validation set.}
\tablestyle{1.5pt}{1.1}
\begin{tabular}{l x{25} x{35}x{35}x{35} x{35}}
\toprule
Method & \begin{tabular}{@{}c@{}}Two \\ frames\end{tabular}  & MOTA $\uparrow$ & IDF1$\uparrow$ & ID Sw. $\downarrow$   & Speed$\uparrow$ \\[.1em]
\shline
baseline~\cite{fairmot} & \xmark & $69.1\%$ & $72.9\%$ &  \cellcolor{Gray2} \textbf{299} &\cellcolor{Gray1} \textbf{25.6} \\
non-local~\cite{nonlocal} & \xmark  & $67.7\%$ & $70.4\%$ & 311 & 16.60 \\ 
CorrNet~\cite{corrnet} & \xmark  & $70.0\%$ & $73.3\%$ & 303 & 22.93 \\ 
SLC ({\textcolor{Maroon}{ours}}) &  \xmark & $70.3\%$ & \cellcolor{Gray2} \bm{$75.8\%$} & \cellcolor{Gray1}\textbf{258} & 20.19 \\ \hline
concat-raw~\cite{centertrack} &  \cmark & $69.3\%$ & $74.1\%$ & 336 & \cellcolor{Gray2} \textbf{23.99}  \\
concat-feat~\cite{ctracker} &  \cmark &$70.4\%$& $74.0\%$ & 308 & 19.77  \\
TLC ({\textcolor{Maroon}{ours}}) &  \cmark &\cellcolor{Gray2}\bm{$70.9\%$} & $74.7\%$ & 326 & 19.26 \\ \hline
STLC ({\textcolor{Maroon}{ours}}) &  \cmark &\cellcolor{Gray1}\bm{$71.5\%$} & \cellcolor{Gray1}\bm{$76.1\%$}  & 307 & 16.56 \\
\bottomrule
\end{tabular}
\label{tab:archablation}
\vspace{-5mm}
\end{table}

\noindent\textbf{Spatial correlation.} 
In order to evaluate the effectiveness of our spatial local correlation module (SLC), we compared our baseline model~\cite{fairmot} and two relation methods, the non-local module~\cite{nonlocal} and CorrNet~\cite{corrnet}. 
Directly use of non-local brings performance degradation on MOT since non-local module does not record the relative positional relationship between the targets and usually brings performance drops on small object~\cite{dnonlocal}. 
In addition, non-local method has a huge overhead in memory and computation. 
In video recognition, a learnable correlation filter network~\cite{corrnet} is proposed, and the grouped convolution is used to reduce the amount of calculation. 
Our method has achieved similar MOTA compared to CorrNet, but has a large improvement on IDF1, which comes from our multi-scale correlation design.
Compared with the baseline, our IDF1 has increased by $2.9\%$, and identity switches have been reduced by $15\%$, demonstrating the discrimination of our spatial correlation.
The Re-ID embeddings cannot easily distinguish similar distractor, our correlation feature models geometric information and is better suited for MOT.

\noindent\textbf{Temporal correlation.} In the field of video object detection~\cite{imagenet}, temporal and global information are commonly used to improve performance~\cite{mega}. 
The research of temporal detection for MOT is still preliminary. 
Recently, CenterTrack~\cite{centertrack} concatenates the previous frame and the current frame, and Chained-Tracker~\cite{ctracker} concatenates the high-level image features for two frames to fuse temporal information. 
We compare these two methods, dubbed \textit{concat-raw} and \textit{concat-feat}, with our temporal local correlation module (TLC).
These two methods brought $0.2\%$ and $1.3\%$ MOTA improvements over the single frame baseline. 
Compared with these two methods, our temporal local correlation module achieves consistent improvements in both MOTA and IDF1. 
Our temporal correlation module helps for the temporal feature alignment around frames. 
At the same time, our method only adds a small overhead to feature-level concatenation, which proves the efficiency of our algorithm.
Compared with the baseline FairMOT, employing both spatial and temporal local correlation (STLC) yields a IDF1 of $76.1\%$ , which brings $3.2\%$ improvement.

\begin{table}[t]
\caption{Ablation studies on MOT17 validation set. ``LT'' and ``Self'' denote using the proposed long-term memory and self-supervised loss, respectively.}

\tablestyle{1.5pt}{1.1}
\begin{tabular}{l x{35}x{35}x{35} x{35}}
\toprule
Method & MOTA $\uparrow$ & IDF1$\uparrow$ & ID Sw. $\downarrow$ & Speed$\uparrow$ \\[.1em]
\shline
STLC & $71.5\%$ & \cellcolor{Gray2}\bm{$76.1\%$}  & 307 & \cellcolor{Gray1}\textbf{16.56}  \\
STLC+LT & \cellcolor{Gray2}\bm{$72.1\%$} & $75.6\%$ & 311 & \cellcolor{Gray2}\textbf{15.62} \\
STLC+LT+Track Loss & \cellcolor{Gray2}\bm{$72.1\%$}  & \cellcolor{Gray2}\bm{$76.1\%$}  & \cellcolor{Gray1}\textbf{299} & \cellcolor{Gray2}\textbf{15.62} \\
STLC+LT+Self Loss & \cellcolor{Gray1}\bm{$72.4\%$}  & \cellcolor{Gray1}\bm{$77.6\%$}  & \cellcolor{Gray2}\textbf{301} & \cellcolor{Gray2}\textbf{15.62} \\
\bottomrule
\end{tabular}
\label{tab:moduleablation}
\vspace{-5mm}
\end{table}

\noindent\textbf{Long-term dependences.} We also analyze the performance improvement with long-term correlation. 
Compared with the method using two frames as the source cues, our long-term method achieves a large improvement of $0.6\%$ MOTA, due to the increased capacity for object detections. The improvement of MOTA also means an increase in the upper bound of our tracker.

\noindent\textbf{Self-supervised learning. } For correlation learning, explicit supervision is usually not imposed~\cite{nonlocal,corrnet}.
We have proposed two supervision methods in section~\ref{sec:self}. 
It can be seen that the siamese tracking supervision imposed to training has achieved a relatively good improvement in IDF1. There is no change in the run time of our algorithm, because the change in training loss does not change the inference processing.
Self-supervised losses have also been appropriately improved on both MOTA and IDF1 due to more positive samples employed in correlation volume.

\begin{figure}[t]
\includegraphics[width=0.5\textwidth]{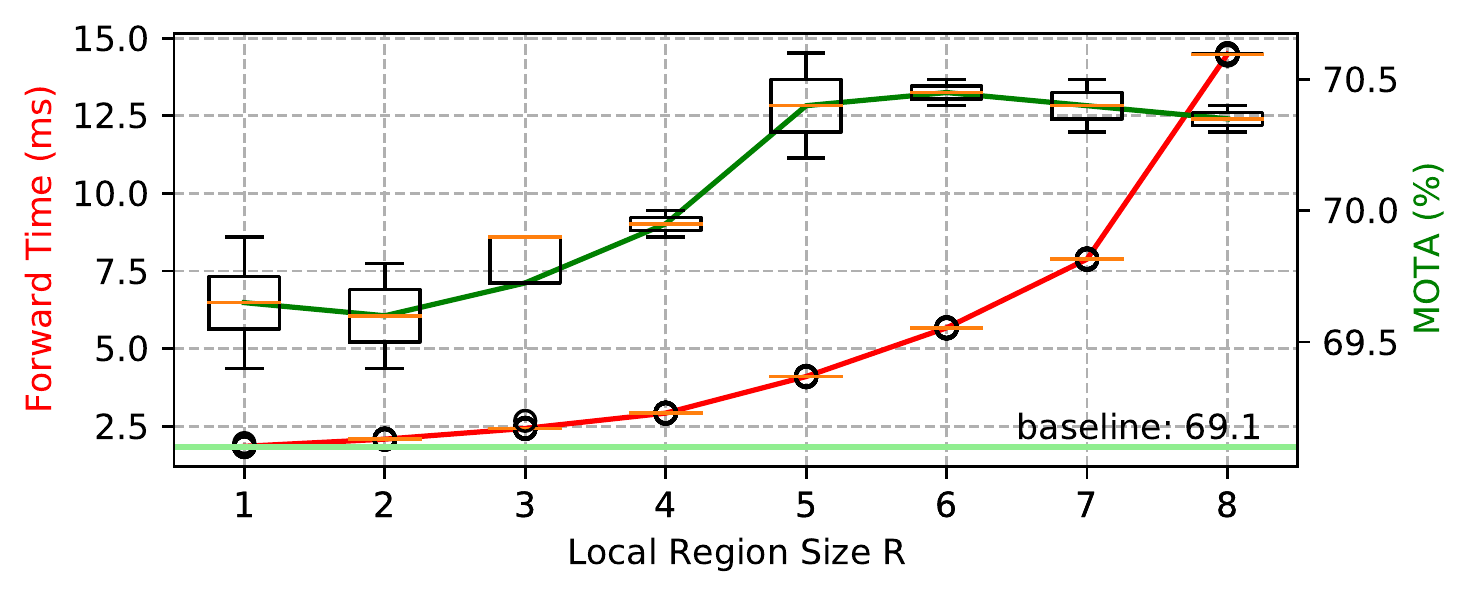}
\caption{ Effect of filter size $R$ on speed and MOTA accuracy on MOT17-val.}
\label{fig:param}
\vspace{-5mm}
\end{figure}

\noindent\textbf{Choice of local region. } Figure~\ref{fig:param} shows the MOTA and run time of our correlation module for different region size $R \in \{1, 2, ..., 8\}$. 
As expected, a larger local size $R$ can cover a larger neighborhood while matching pixels, thus yields a higher accuracy. 
But the improvements become marginal beyond $R = 5$, possibly due to the low resolution of the feature maps.
Note that non-local module usually doubles the run time of the backbone and the cost of explicitly computing optical flow~\cite{flownet} can be very high as well. 
This shows that our correlation module is more efficient by learning motion information from features directly.
Region size $R=5$ yields a good trade-off between speed and accuracy.
The computation overhead is relatively small, compared to the complexity of the whole detection networks.

\subsection{Experiments on MOT Challenges} 
\label{sec:sota}

\begin{table}[t]
\caption{Comparisons of our method with state-of-the-arts on MOT benchmarks~\cite{mot15,mot16,mot20}. We set new state-of-the-art results by a significant margin in terms of MOTA and IDF1. Our correlation tracker is more accurate while running with high speed.}
\vspace{-5mm}
\center
\tabcolsep=0.11cm
\resizebox{\columnwidth}{!}{
\begin{tabular}{l c c c c c c c c c }
\toprule
Method & MOTA $\uparrow$ & IDF1 $\uparrow$ & MT $\uparrow$ & ML $\downarrow$ & FP $\downarrow$ & FN $\downarrow$ & ID Sw. $\downarrow$  & Hz  $\uparrow$\\ [0.5ex] 
\midrule
\multicolumn{8}{c}{2D MOT 2015~\cite{mot15}    } \\
\midrule
\textbf{CorrTracker} ({\textcolor{Maroon}{ours}}) & \textbf{62.3} &  \textbf{65.7} & \textbf{49.0} &  12.9  & 6909 & \textbf{15728} & 513 & 17.9 \\
JDMOTGNN~\cite{jdmotgnn} & 60.7 & 64.6 & 47.0 & \textbf{10.5} & 7334 & 16358 & 477 & 2.4   \\
FairMOTv2~\cite{fairmot} & 60.6 & 64.7 & 47.6 & 11.0 & 7854 & 15785 & 591 & \textbf{30.5}   \\
Tube\_TK\_POI~\cite{tubetk}  & 58.4 & 53.1 &  39.3 & 18.0 & 5756 &  18961 &  854 & 5.8 \\
MPN~ \cite{mpn} & 51.5 &  58.6 & 31.2 &  25.9  & 7620& 21780 & \textbf{375} & 6.5\\
Tracktor++v2~ \cite{tracktor} & 46.6 &  47.6 & 18.2 &  27.9  & \textbf{4624} & 26896 & 1290 & 1.8\\
\midrule
\multicolumn{8}{c}{MOT16~\cite{mot16}} \\
 \midrule
\textbf{CorrTracker} ({\textcolor{Maroon}{ours}}) &  \textbf{76.6} & \textbf{74.3} & \textbf{47.8} &  \textbf{13.3} & 10860 & \textbf{30756} & 979 & 14.8 \\
FairMOTv2~\cite{fairmot} & 74.9 & 72.8 & 44.7 & 15.9 & 10163 &  34484 & 1074 & \textbf{25.4}  \\
CTracker~\cite{ctracker}  & 67.6 & 57.2 &  32.9 & 19.0 & 8934 &  48305 &  1897 & 6.8 \\
Tube\_TK\_POI~\cite{tubetk}  & 66.9 & 62.2 &  39.0 & 16.1 & 11544 &  47502 &  1236 & 1.0 \\
POI~\cite{poi} & 66.1 & 65.1 &  34.0 & 20.8 & 5061 &  55914 &  805 & 9.9 \\
Tracktor++v2~\cite{tracktor} & 56.2 &  54.9  & 20.7 & 35.8 &  \textbf{2394} & 76844 & \textbf{617} & 1.8\\
 \midrule
\multicolumn{8}{c}{ MOT17~\cite{mot16}   } \\
\midrule
\textbf{CorrTracker} ({\textcolor{Maroon}{ours}}) & \textbf{76.5} & \textbf{73.6} & \textbf{47.6} & \textbf{12.7} & 29808 & \textbf{99510} & 3369 & 14.8 \\
FairMOTv2~\cite{fairmot} & 73.7 &  72.3 & 43.2 & 17.3 & 27507 & 117477 & 3303 & \textbf{ 25.9}  \\
CenterTrack~\cite{centertrack} & 67.8 & 64.7 &  34.6 & 24.6 & 18498 &  160,332 & 3039 & 3.8 \\
CTracker~\cite{ctracker}  & 66.6 & 57.4 &  32.2 & 24.2 & 22284 &  160491 &  5529 & 6.8 \\
Tracktor++v2~\cite{tracktor}     & 56.3 & 55.1 & 21.1 & 35.3 & \textbf{8866} & 235449 & \textbf{1987} & 1.8\\
 \midrule
\multicolumn{8}{c}{ MOT20~\cite{mot20}   } \\
\midrule
\textbf{CorrTracker} ({\textcolor{Maroon}{ours}}) & 65.2 & \textbf{69.1} & 66.4 & 8.9 & 79429 & 95855 & 5183 & 8.5 \\
JDMOTGNN~\cite{jdmotgnn} & \textbf{67.1} &  67.5 & 53.1 & 13.2 & \textbf{31913} & 135409 & \textbf{3131} & 0.9  \\
FairMOTv2~\cite{fairmot} & 61.8 &  67.3 & \textbf{68.8} & \textbf{7.6} & 103440 & \textbf{88901} & 5243 &\textbf{13.2}  \\
\bottomrule
\vspace{-10mm}
\end{tabular}}

\label{tab:mot}
\end{table}

To extensively evaluate the proposed method, we compare it with 8 state-of-the-art trackers, which cover most of current representative methods. 
There are 2 joint detection and embedding methods (JDE~\cite{jde} and FairMOT~\cite{fairmot}), 2 multi-frame prediction methods (Tube{\_}TK~\cite{tubetk} and CTracker~\cite{ctracker}, 2 graph network based methods (MPN~\cite{mpn} and JDMOTGNN~\cite{jdmotgnn}), 2 offset prediction based methods (CenterTracker~\cite{centertrack} and Tracktor++v2~\cite{tracktor}). 
The results are summarized in Table~\ref{tab:mot}.

\textbf{2DMOT2015~\cite{mot15}.} The evaluation on 2DMOT2015 is performed by the official toolkit.
As shown in Table \ref{tab:mot}, our correlation tracker outperforms the top private method of 2DMOT2015, (\emph{i.e.}, FairMOT~\cite{fairmot}), by $1.7\%$ in MOTA and $1.0\%$ in IDF1. 
It is worth noting that the ID Switches are decreased by $13\%$, which shows the robustness of our correlation module.
Moreover, our tracker is superior to the recent end-to-end graph trackers JDMOTGNN~\cite{jdmotgnn}. 
Our feature propagation approach can absorb both foreground and background information, which improves our tracker by $1.6\%$ in terms of MOTA. 

\textbf{MOT16~\cite{mot16} and MOT17~\cite{mot16}.} Table~\ref{tab:mot} reports the evaluation results with the comparisons to recent prevailing trackers on MOT16. 
The recent proposed FairMOTv2~\cite{fairmot} achieves the second performance in MOTA and IDF1, while our method ranks first with $76.6\%$ MOTA, outperforming other private approaches by a significant margin.
Our CorrTracker achieves the best performance on MOT17, surpassing FairMOTv2 by $2.8\%$ MOTA and $1.3\%$ IDF1. 
Moreover, the FN of our CorrTracker surpasses FairMOTv2 by $15\%$, which means nearly 20,000 new bounding boxes are added to the association process. In this case, our algorithm still maintains comparable or even superior ID Switches, which actually proves that our method significantly improves the tracking association.
As reported in Table~\ref{tab:mot}, our CorrTracker, CenterTrack~\cite{centertrack} and CTrack~\cite{ctracker} all use multi-frame cues to predict detections, our FN is largely decreased by $30\%$.

\textbf{MOT20~\cite{mot20}.} To further evaluate the proposed models, we report the results on MOT20, which is more challenging than MOT17. The final results is presented in the bottom block of Table~\ref{tab:mot}. 
Our CorrTracker achieves MOTA score of $65.2\%$, substantially outperforming FairMOTv2~\cite{fairmot} with MOTA of $61.8\%$. Although our approach is an order of magnitude faster than JDMOTGNN~\cite{jdmotgnn} in speed, our accuracy is slightly worse due to the anchor-free design.

\begin{figure*}[t]
\center
\includegraphics[width=0.99\textwidth]{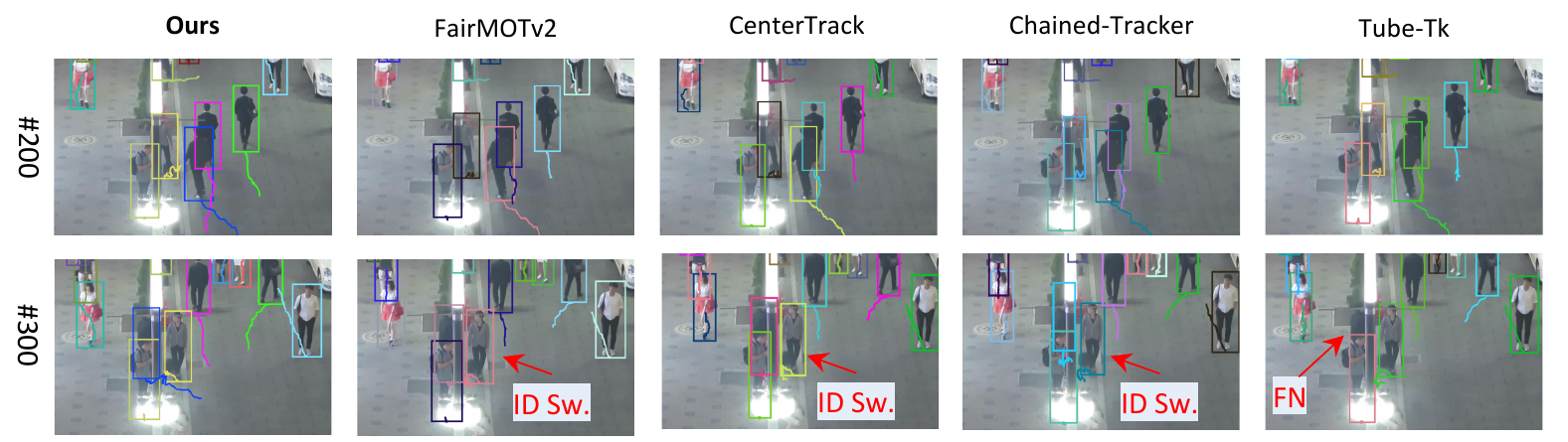}
\caption{Qualitative comparisons against several prior methods~\cite{fairmot,centertrack,ctracker, tubetk} in occlusion situations. Frames are sampled from MOT17-03. Our CorrTracker can identify objects via mining the context patterns around targets. }
\label{fig:cmp}
\end{figure*}
\begin{figure*}[t]
\centering
\setlength{\tabcolsep}{0.25ex}

\begin{tabular}
{c cc c cc}
\mbox{\rotatebox[x=-0.5cm]{90}{\small{MOT17-01}}}
&\includegraphics[width = 0.23\textwidth]{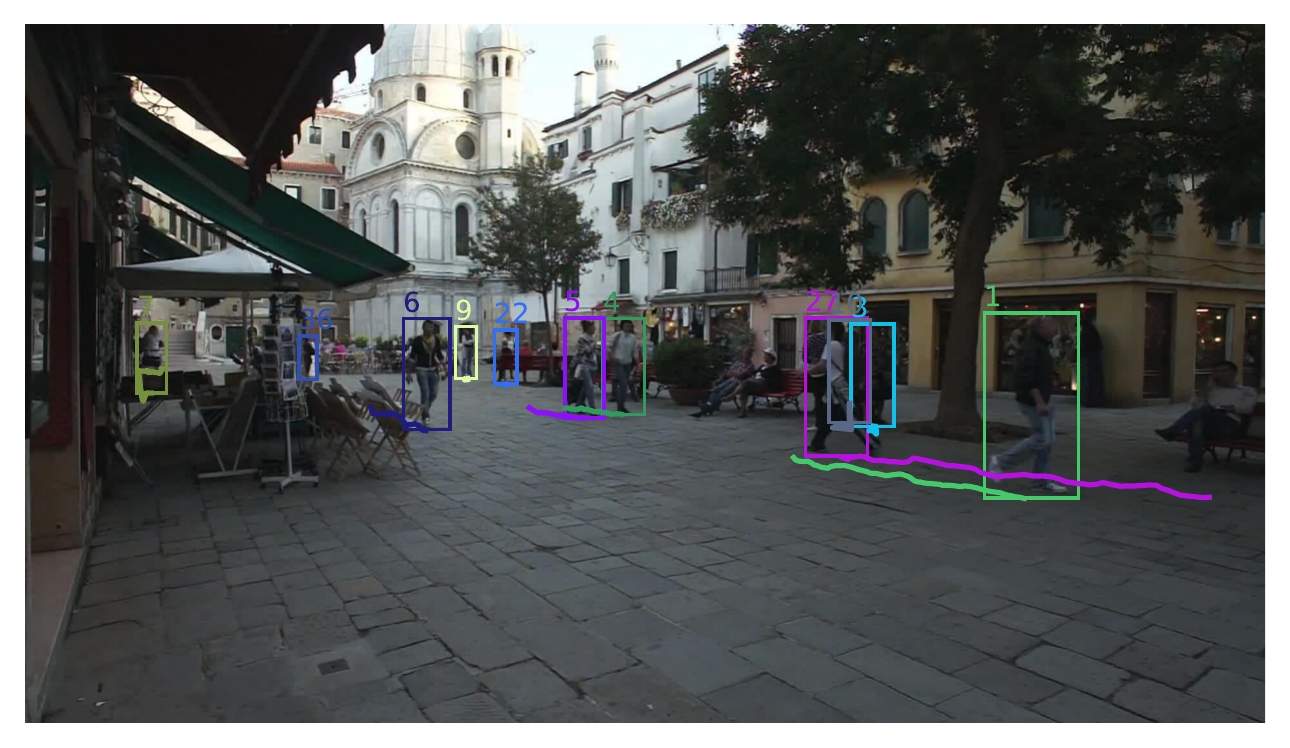}
&\includegraphics[width = 0.23\textwidth]{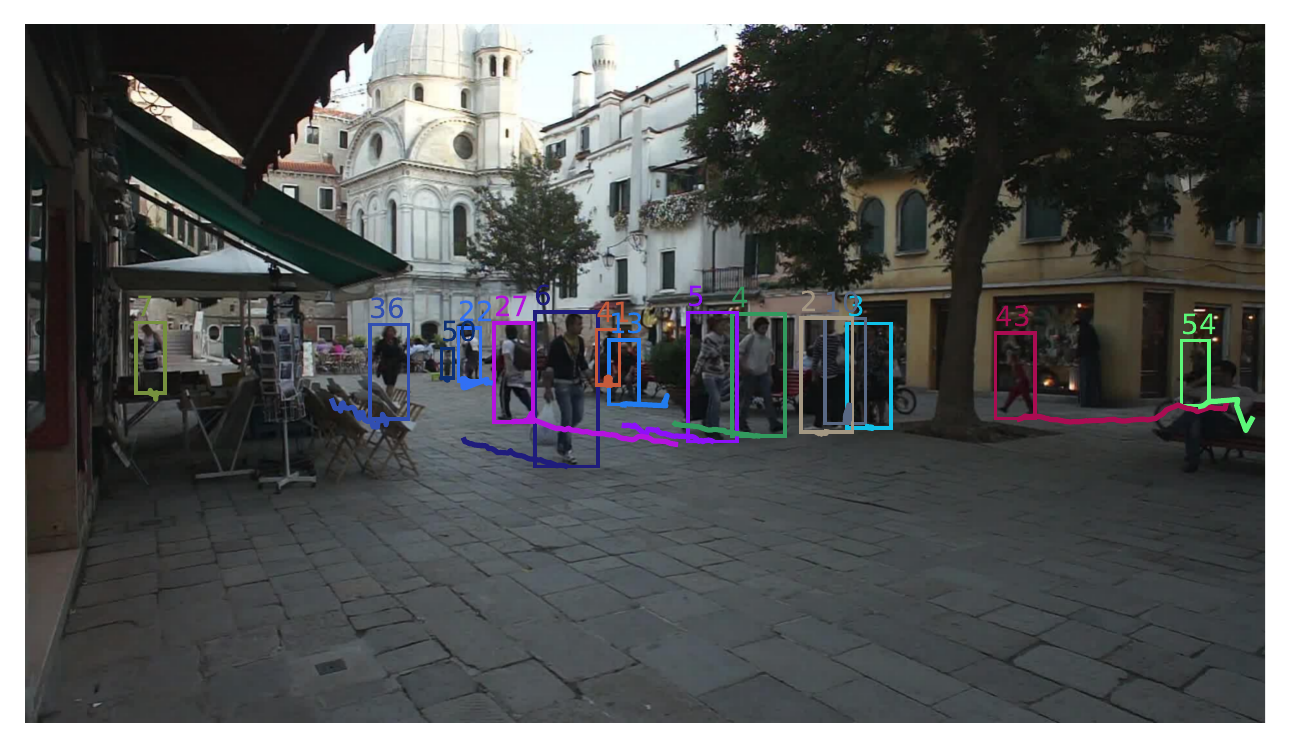}
&\mbox{\rotatebox[x=-0.5cm]{90}{\small{MOT17-07}}}
&\includegraphics[width = 0.23\textwidth]{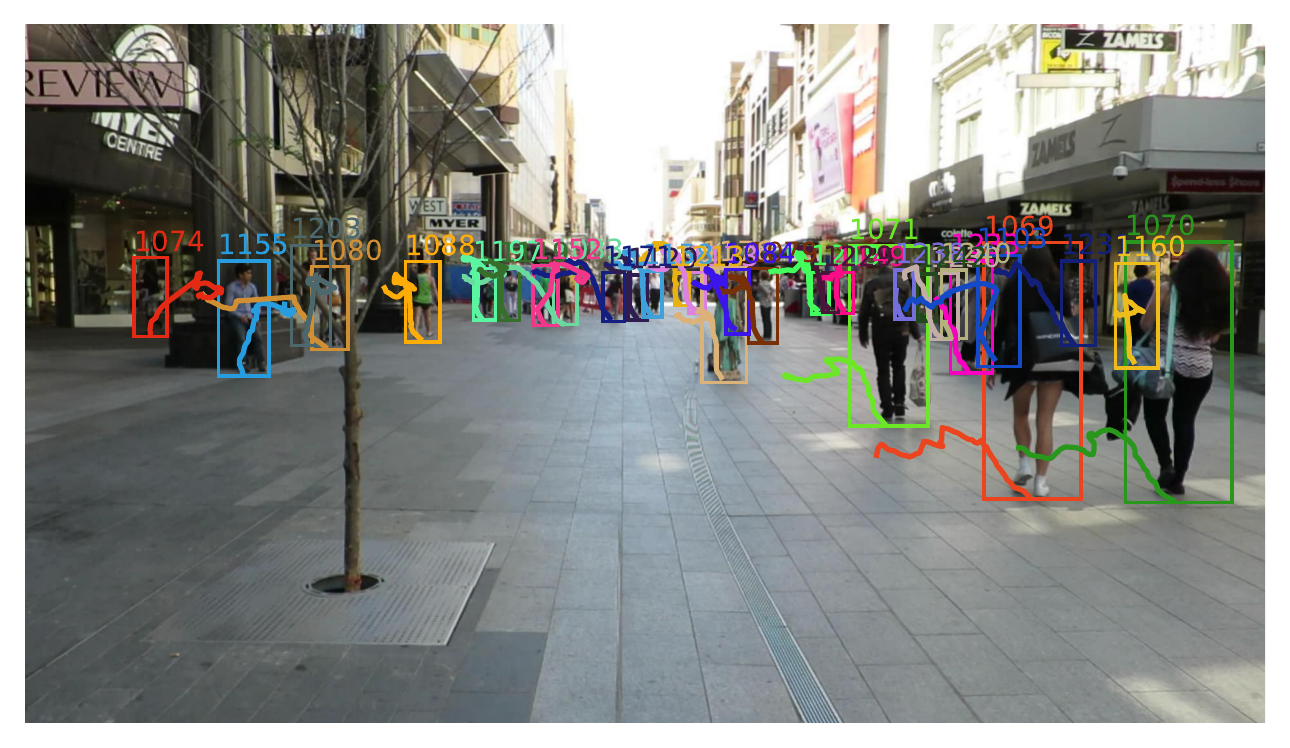}
&\includegraphics[width = 0.23\textwidth]{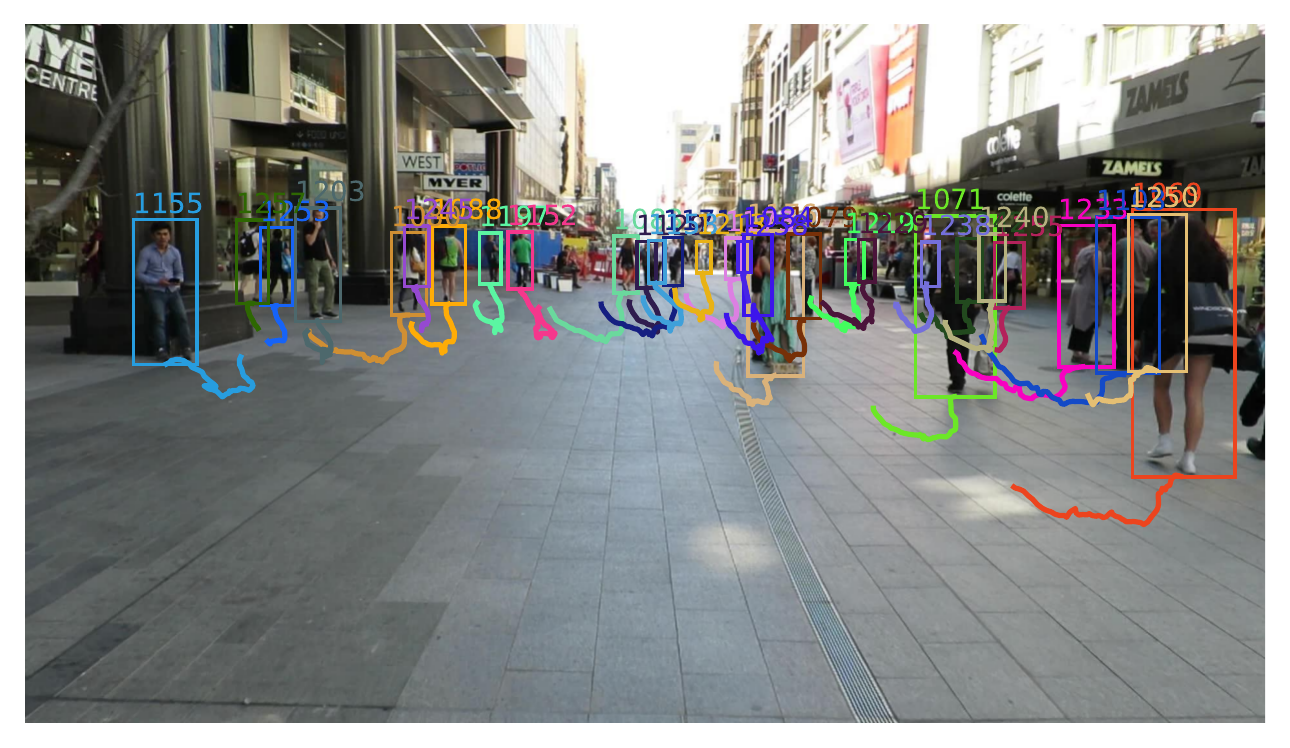}
\\
\mbox{\rotatebox[x=-0.5cm]{90}{\small{MOT17-12}}}
&\includegraphics[width = 0.23\textwidth]{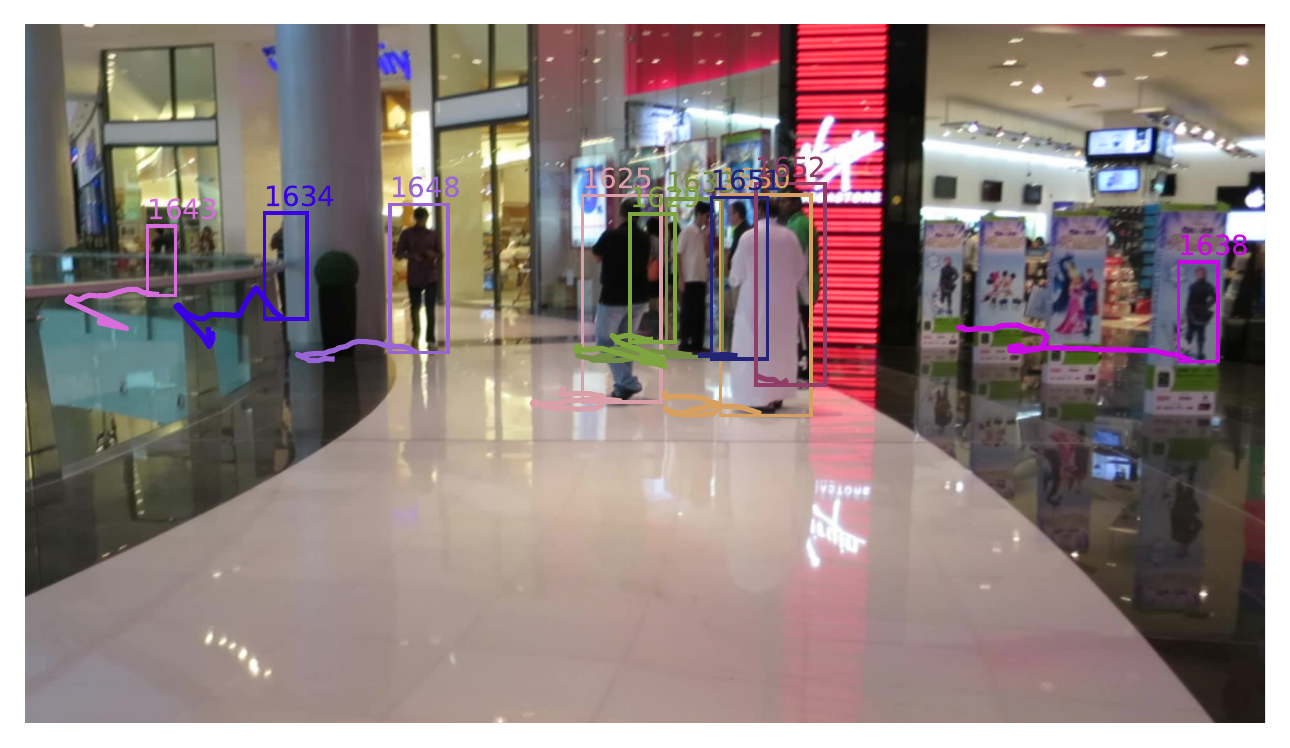}
&\includegraphics[width = 0.23\textwidth]{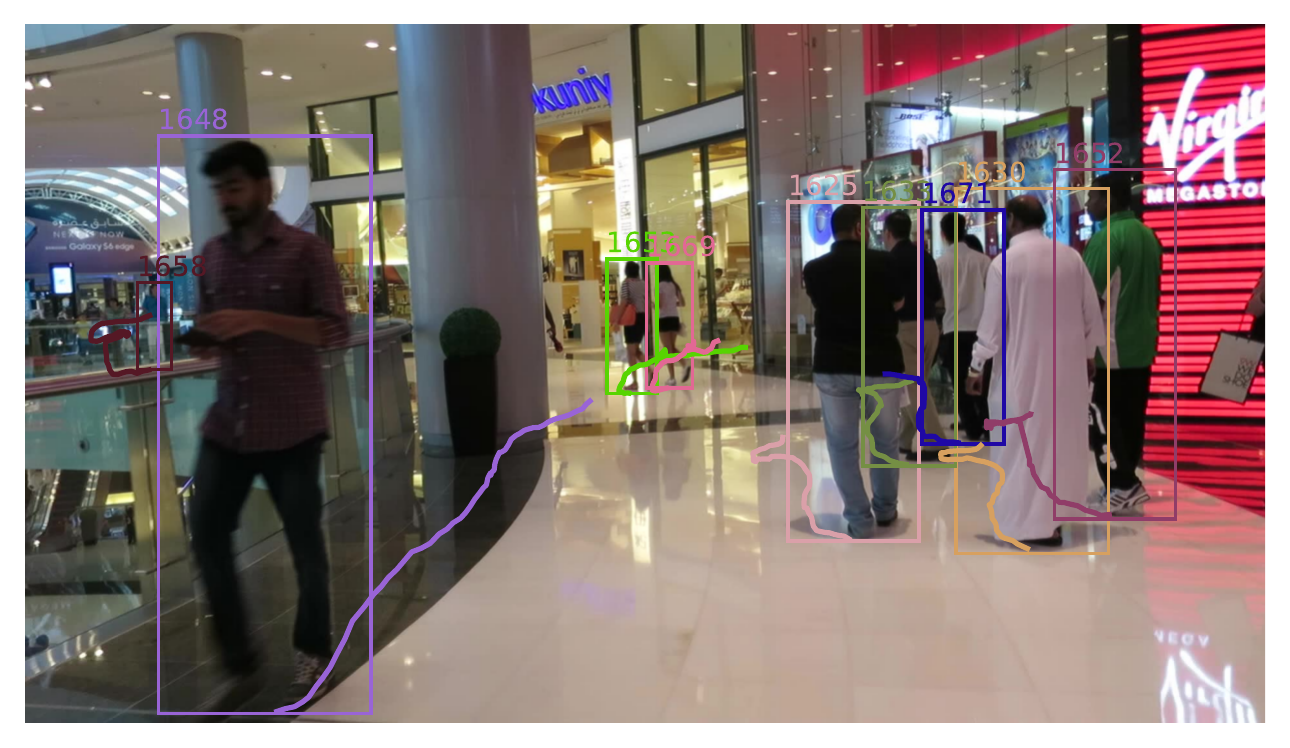}
&\mbox{\rotatebox[x=-0.5cm]{90}{\small{MOT17-14}}}
&\includegraphics[width = 0.23\textwidth]{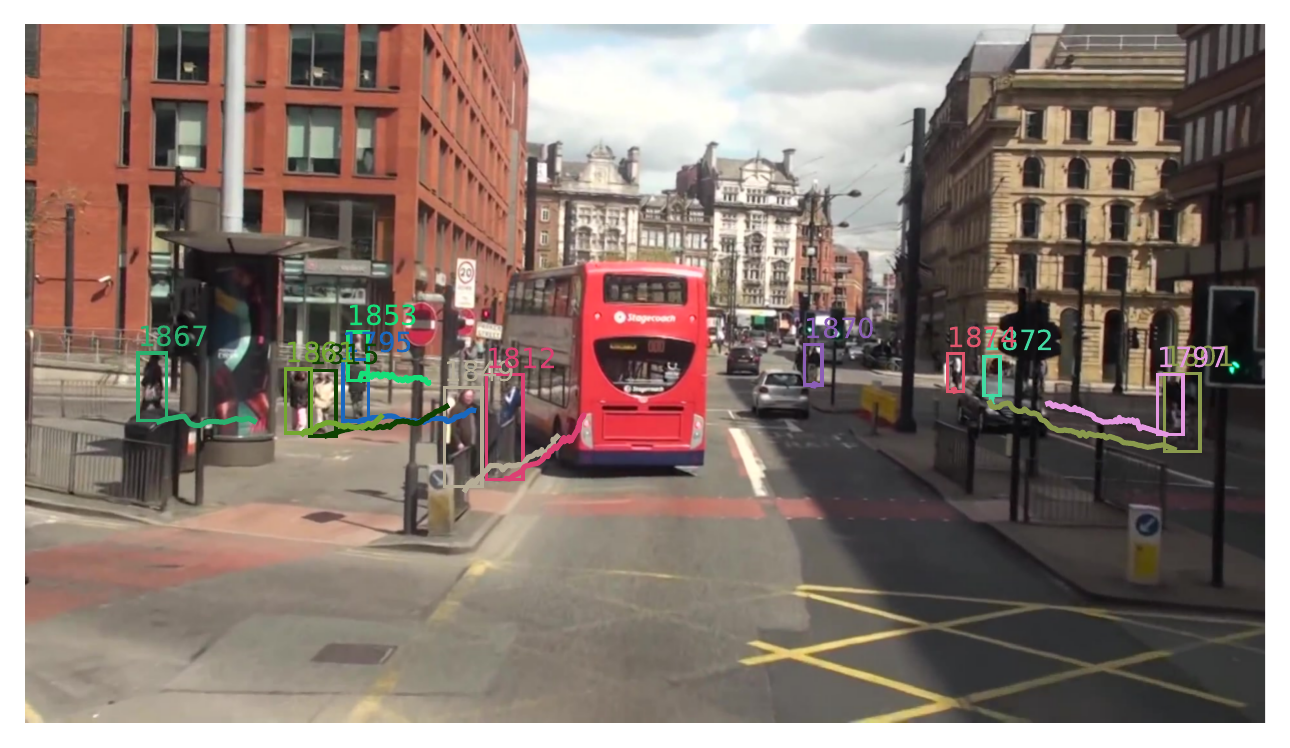}
&\includegraphics[width = 0.23\textwidth]{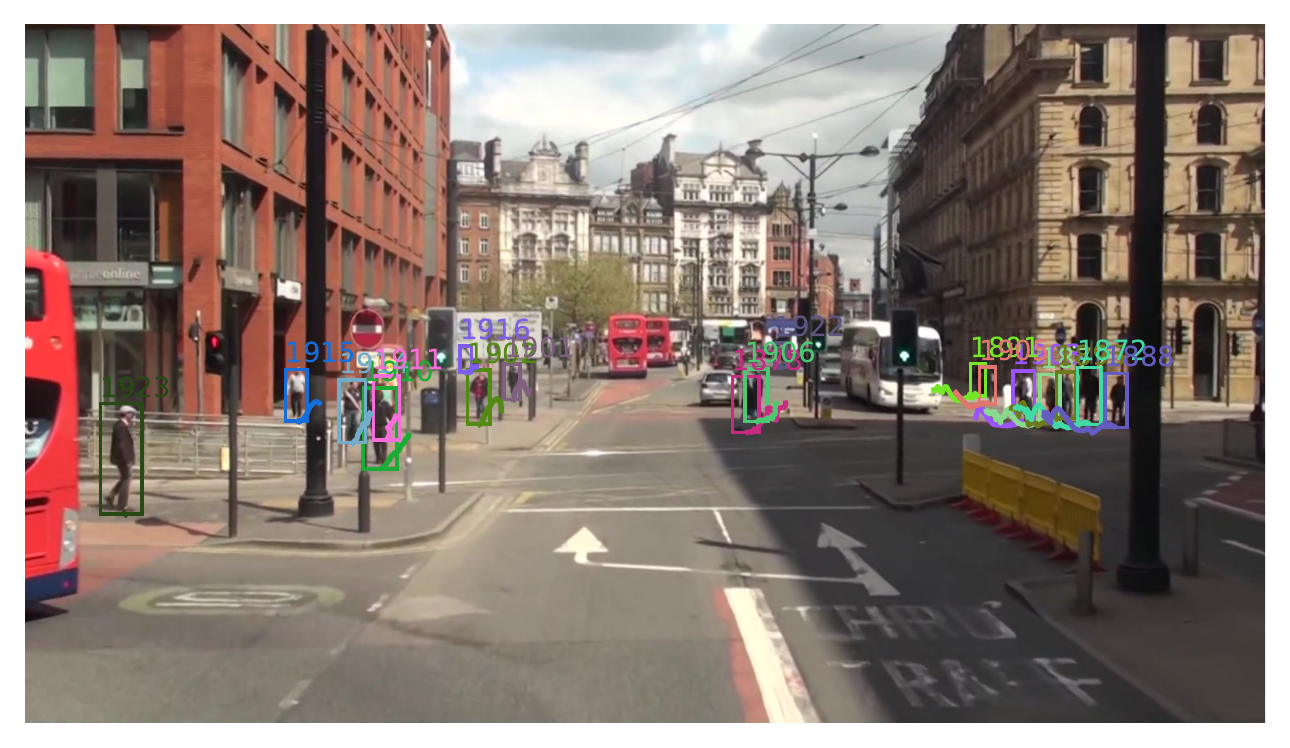}
\\[-5mm]
\mbox{\rotatebox[x=-1cm]{90}{\small{MOT20-06}}}
&\multicolumn{2}{c}{\includegraphics[width = 0.49\textwidth]{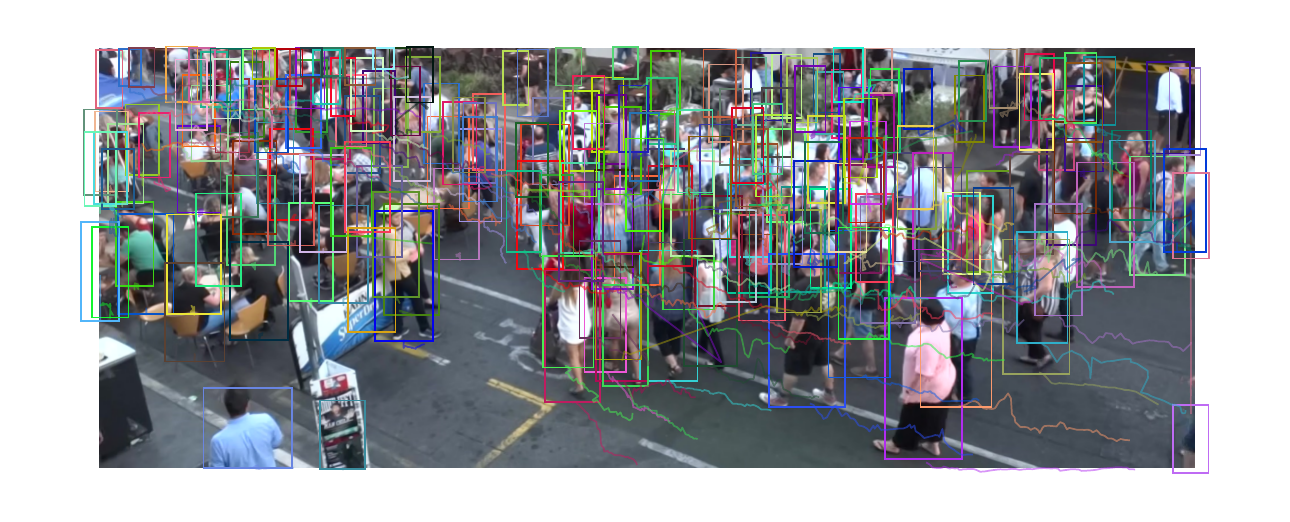}}
&\mbox{\rotatebox[x=-1cm]{90}{\small{MOT20-08}}}
&\multicolumn{2}{c}{\includegraphics[width = 0.48\textwidth]{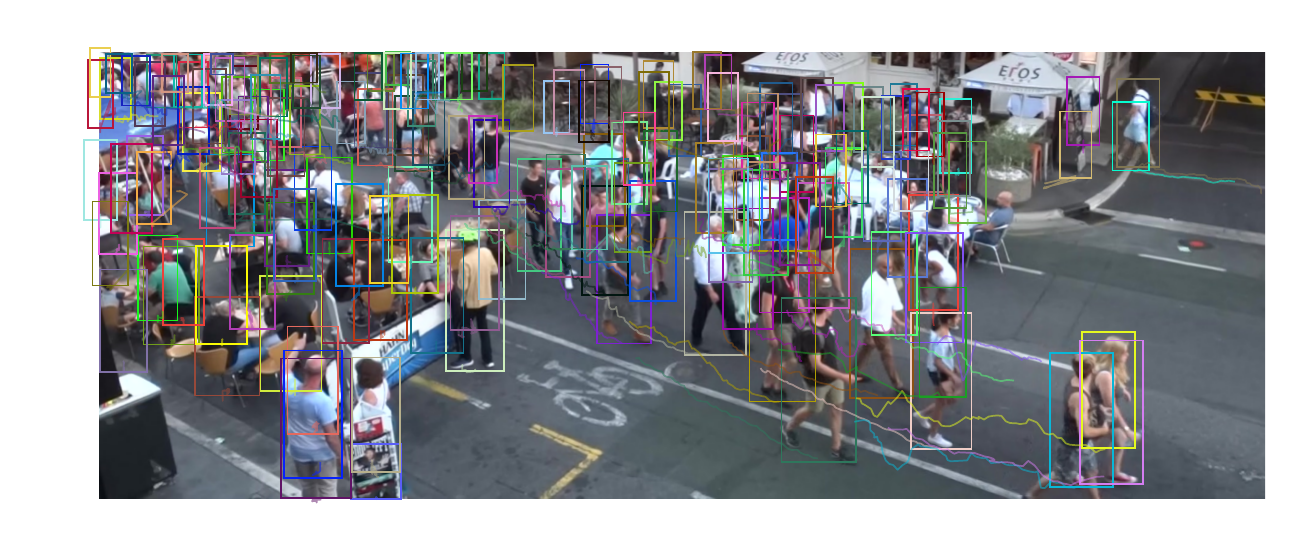}}
\\
\end{tabular}

\caption{Qualitative results of our correlation tracker on MOT17~\cite{mot16} and MOT20~\cite{mot20}.  The color of each bounding box indicates the target identity. The dotted line under each bounding box denotes the recent tracklet of each target. The proposed tracker predicts trajectories with substantially robust and temporally consistent.}
\label{fig:mot}
\end{figure*}

\subsection{Visualization}
\label{sec:vis}

We visualize the tracking trajectories for prior methods, \textit{i.e.}, the center offset~\cite{centertrack} and multi-frame bounding boxes~\cite{ctracker,tubetk}, in Figure~\ref{fig:cmp}. 
We observe that our correlation map focuses on the entire context, while the regular appearance feature concentrates on the local region of the target. 
Our correlation module improves the reliability of recognition since it provides a global view of the target. 
Methods based on offset prediction, \emph{e.g.}, CenterTrack~\cite{centertrack} and CTracker~\cite{ctracker}, can easily generate id switches when encountered with complex object interactions.
Figure ~\ref{fig:mot} shows qualitative results of our Correlation Tracker on MOT17 and MOT20, the advantage over existing method is most pronounced on the robust to occlusion and tiny objects.

\section{Conclusion}
\label{sec:conclusion}

In this work, we propose a novel correlation tracking framework based upon the observation that the relational structure helps to distinguish similar objects.
Our correlation module densely matches all targets with their local context and learn a discriminative embeddings from the correlation volumes.
Furthermore, we show how to extend the correlation module from spatial layout to the adjacent frames for strengthening the temporal modeling ability.
We explore that self-supervised learning to impose a discriminative constraint on the correlation volume, which explicitly predicts a instance flow.
Extensive experiments on four MOT challenges demonstrate that our CorrTracker achieves state-of-the-art performance and is efficient in inference.

{
\newpage
\small
\bibliographystyle{ieee_fullname}
\bibliography{egbib}

\begin{thebibliography}{10}\itemsep=-1pt

\bibitem{tracktor}
Philipp Bergmann, Tim Meinhardt, and Laura Leal-Taixe.
\newblock Tracking without bells and whistles.
\newblock In {\em Proceedings of the IEEE international conference on computer
  vision}, pages 941--951, 2019.

\bibitem{siamfc}
Luca Bertinetto, Jack Valmadre, Joao~F Henriques, Andrea Vedaldi, and Philip~HS
  Torr.
\newblock Fully-convolutional siamese networks for object tracking.
\newblock In {\em European conference on computer vision}, pages 850--865.
  Springer, 2016.

\bibitem{sort}
Alex Bewley, Zongyuan Ge, Lionel Ott, Fabio Ramos, and Ben Upcroft.
\newblock Simple online and realtime tracking.
\newblock In {\em 2016 IEEE International Conference on Image Processing
  (ICIP)}, pages 3464--3468. IEEE, 2016.

\bibitem{ioutracker}
Erik Bochinski, Volker Eiselein, and Thomas Sikora.
\newblock High-speed tracking-by-detection without using image information.
\newblock In {\em 2017 14th IEEE International Conference on Advanced Video and
  Signal Based Surveillance (AVSS)}, pages 1--6. IEEE.

\bibitem{mpn}
Guillem Bras{\'o} and Laura Leal-Taix{\'e}.
\newblock Learning a neural solver for multiple object tracking.
\newblock In {\em Proceedings of the IEEE/CVF Conference on Computer Vision and
  Pattern Recognition}, pages 6247--6257, 2020.

\bibitem{mega}
Yihong Chen, Yue Cao, Han Hu, and Liwei Wang.
\newblock Memory enhanced global-local aggregation for video object detection.
\newblock In {\em Proceedings of the IEEE/CVF Conference on Computer Vision and
  Pattern Recognition}, pages 10337--10346, 2020.

\bibitem{famnet}
Peng Chu and Haibin Ling.
\newblock Famnet: Joint learning of feature, affinity and multi-dimensional
  assignment for online multiple object tracking.
\newblock In {\em Proceedings of the IEEE International Conference on Computer
  Vision}, pages 6172--6181, 2019.

\bibitem{mot20}
P. Dendorfer, H. Rezatofighi, A. Milan, J. Shi, D. Cremers, I. Reid, S. Roth,
  K. Schindler, and L. Leal-Taix\'{e}.
\newblock Mot20: A benchmark for multi object tracking in crowded scenes.
\newblock {\em arXiv:2003.09003[cs]}, Mar. 2020.
\newblock arXiv: 2003.09003.

\bibitem{caltech}
Piotr Doll{\'a}r, Christian Wojek, Bernt Schiele, and Pietro Perona.
\newblock Pedestrian detection: A benchmark.
\newblock In {\em 2009 IEEE Conference on Computer Vision and Pattern
  Recognition}, pages 304--311. IEEE, 2009.

\bibitem{flownet}
Alexey Dosovitskiy, Philipp Fischer, Eddy Ilg, Philip Hausser, Caner Hazirbas,
  Vladimir Golkov, Patrick Van Der~Smagt, Daniel Cremers, and Thomas Brox.
\newblock Flownet: Learning optical flow with convolutional networks.
\newblock In {\em Proceedings of the IEEE international conference on computer
  vision}, pages 2758--2766, 2015.

\bibitem{eth}
Andreas Ess, Bastian Leibe, Konrad Schindler, and Luc Van~Gool.
\newblock A mobile vision system for robust multi-person tracking.
\newblock In {\em 2008 IEEE Conference on Computer Vision and Pattern
  Recognition}, pages 1--8. IEEE, 2008.

\bibitem{dt}
Christoph Feichtenhofer, Axel Pinz, and Andrew Zisserman.
\newblock Detect to track and track to detect.
\newblock In {\em Proceedings of the IEEE International Conference on Computer
  Vision}, pages 3038--3046, 2017.

\bibitem{dpm}
Pedro~F Felzenszwalb, Ross~B Girshick, David McAllester, and Deva Ramanan.
\newblock Object detection with discriminatively trained part-based models.
\newblock {\em IEEE transactions on pattern analysis and machine intelligence},
  32(9):1627--1645, 2009.

\bibitem{fastrcnn}
Ross Girshick.
\newblock Fast r-cnn.
\newblock In {\em Proceedings of the IEEE international conference on computer
  vision}, pages 1440--1448, 2015.

\bibitem{tmlf}
Roberto Henschel, Laura Leal-Taix{\'e}, Bodo Rosenhahn, and Konrad Schindler.
\newblock Tracking with multi-level features.
\newblock {\em arXiv preprint arXiv:1607.07304}, 2016.

\bibitem{bjd}
Roberto Henschel, Yunzhe Zou, and Bodo Rosenhahn.
\newblock Multiple people tracking using body and joint detections.
\newblock In {\em Proceedings of the IEEE Conference on Computer Vision and
  Pattern Recognition Workshops}, pages 0--0, 2019.

\bibitem{lift}
Andrea Hornakova, Roberto Henschel, Bodo Rosenhahn, and Paul Swoboda.
\newblock Lifted disjoint paths with application in multiple object tracking.
\newblock In {\em The 37th International Conference on Machine Learning
  (ICML)}, July 2020.

\bibitem{relationnet}
Han Hu, Jiayuan Gu, Zheng Zhang, Jifeng Dai, and Yichen Wei.
\newblock Relation networks for object detection.
\newblock In {\em Proceedings of the IEEE Conference on Computer Vision and
  Pattern Recognition}, pages 3588--3597, 2018.

\bibitem{autodrive}
Joel Janai, Fatma G{\"u}ney, Aseem Behl, Andreas Geiger, et~al.
\newblock Computer vision for autonomous vehicles: Problems, datasets and state
  of the art.
\newblock {\em Foundations and Trends{\textregistered} in Computer Graphics and
  Vision}, 12(1--3):1--308, 2020.

\bibitem{kalman}
R.~E. Kalman.
\newblock A new approach to linear filtering and prediction problems.
\newblock {\em ASME Journal of Basic Engineering}, 1960.

\bibitem{clear}
Rangachar Kasturi, Dmitry Goldgof, Padmanabhan Soundararajan, Vasant Manohar,
  John Garofolo, Rachel Bowers, Matthew Boonstra, Valentina Korzhova, and Jing
  Zhang.
\newblock Framework for performance evaluation of face, text, and vehicle
  detection and tracking in video: Data, metrics, and protocol.
\newblock {\em IEEE Transactions on Pattern Analysis and Machine Intelligence},
  31(2):319--336, 2008.

\bibitem{adam}
Diederik~P Kingma and Jimmy Ba.
\newblock Adam: A method for stochastic optimization.
\newblock {\em arXiv preprint arXiv:1412.6980}, 2014.

\bibitem{hungarian}
Harold~W Kuhn.
\newblock The hungarian method for the assignment problem.
\newblock {\em Naval research logistics quarterly}, 2(1-2):83--97, 1955.

\bibitem{mot15}
L. Leal-Taix\'{e}, A. Milan, I. Reid, S. Roth, and K. Schindler.
\newblock {MOTC}hallenge 2015: {T}owards a benchmark for multi-target tracking.
\newblock {\em arXiv:1504.01942 [cs]}, Apr. 2015.
\newblock arXiv: 1504.01942.

\bibitem{ids}
Yuan Li, Chang Huang, and Ram Nevatia.
\newblock Learning to associate: Hybridboosted multi-target tracker for crowded
  scene.
\newblock In {\em 2009 IEEE Conference on Computer Vision and Pattern
  Recognition}, pages 2953--2960. IEEE, 2009.

\bibitem{coco}
Tsung-Yi Lin, Michael Maire, Serge Belongie, James Hays, Pietro Perona, Deva
  Ramanan, Piotr Doll{\'a}r, and C~Lawrence Zitnick.
\newblock Microsoft coco: Common objects in context.
\newblock In {\em European conference on computer vision}, pages 740--755.
  Springer, 2014.

\bibitem{motdt}
Chen Long, Ai Haizhou, Zhuang Zijie, and Shang Chong.
\newblock Real-time multiple people tracking with deeply learned candidate
  selection and person re-identification.
\newblock In {\em ICME}, 2018.

\bibitem{cosnet}
Xiankai Lu, Wenguan Wang, Chao Ma, Jianbing Shen, Ling Shao, and Fatih Porikli.
\newblock See more, know more: Unsupervised video object segmentation with
  co-attention siamese networks.
\newblock In {\em Proceedings of the IEEE/CVF Conference on Computer Vision and
  Pattern Recognition}, pages 3623--3632, 2019.

\bibitem{retinatrack}
Zhichao Lu, Vivek Rathod, Ronny Votel, and Jonathan Huang.
\newblock Retinatrack: Online single stage joint detection and tracking.
\newblock In {\em Proceedings of the IEEE/CVF Conference on Computer Vision and
  Pattern Recognition}, pages 14668--14678, 2020.

\bibitem{mot16}
A. Milan, L. Leal-Taix\'{e}, I. Reid, S. Roth, and K. Schindler.
\newblock {MOT}16: {A} benchmark for multi-object tracking.
\newblock {\em arXiv:1603.00831 [cs]}, Mar. 2016.
\newblock arXiv: 1603.00831.

\bibitem{surveillance}
Sangmin Oh, Anthony Hoogs, Amitha Perera, Naresh Cuntoor, Chia-Chih Chen,
  Jong~Taek Lee, Saurajit Mukherjee, JK Aggarwal, Hyungtae Lee, Larry Davis,
  et~al.
\newblock A large-scale benchmark dataset for event recognition in surveillance
  video.
\newblock In {\em CVPR 2011}, pages 3153--3160. IEEE, 2011.

\bibitem{tubetk}
Bo Pang, Yizhuo Li, Yifan Zhang, Muchen Li, and Cewu Lu.
\newblock Tubetk: Adopting tubes to track multi-object in a one-step training
  model.
\newblock In {\em Proceedings of the IEEE/CVF Conference on Computer Vision and
  Pattern Recognition}, pages 6308--6318, 2020.

\bibitem{imgtf}
Niki Parmar, Ashish Vaswani, Jakob Uszkoreit, Lukasz Kaiser, Noam Shazeer,
  Alexander Ku, and Dustin Tran.
\newblock Image transformer.
\newblock In {\em Proceedings of the 35th International Conference on Machine
  Learning}, pages 4055--4064. PMLR, 2018.

\bibitem{ctracker}
Jinlong Peng, Changan Wang, Fangbin Wan, Yang Wu, Yabiao Wang, Ying Tai,
  Chengjie Wang, Jilin Li, Feiyue Huang, and Yanwei Fu.
\newblock Chained-tracker: Chaining paired attentive regression results for
  end-to-end joint multiple-object detection and tracking.
\newblock In {\em Proceedings of the European Conference on Computer Vision},
  2020.

\bibitem{yolov3}
Joseph Redmon and Ali Farhadi.
\newblock Yolov3: An incremental improvement.
\newblock {\em arXiv preprint arXiv:1804.02767}, 2018.

\bibitem{fasterrcnn}
Shaoqing Ren, Kaiming He, Ross Girshick, and Jian Sun.
\newblock Faster r-cnn: Towards real-time object detection with region proposal
  networks.
\newblock In {\em Advances in neural information processing systems}, pages
  91--99, 2015.

\bibitem{idf1}
Ergys Ristani, Francesco Solera, Roger Zou, Rita Cucchiara, and Carlo Tomasi.
\newblock Performance measures and a data set for multi-target, multi-camera
  tracking.
\newblock In {\em European Conference on Computer Vision}, pages 17--35.
  Springer, 2016.

\bibitem{imagenet}
Olga Russakovsky, Jia Deng, Hao Su, Jonathan Krause, Sanjeev Satheesh, Sean Ma,
  Zhiheng Huang, Andrej Karpathy, Aditya Khosla, Michael Bernstein, et~al.
\newblock Imagenet large scale visual recognition challenge.
\newblock {\em International journal of computer vision}, 115(3):211--252,
  2015.

\bibitem{crowdhuman}
Shuai Shao, Zijian Zhao, Boxun Li, Tete Xiao, Gang Yu, Xiangyu Zhang, and Jian
  Sun.
\newblock Crowdhuman: A benchmark for detecting human in a crowd.
\newblock {\em arXiv preprint arXiv:1805.00123}, 2018.

\bibitem{pwc}
Deqing Sun, Xiaodong Yang, Ming-Yu Liu, and Jan Kautz.
\newblock Pwc-net: Cnns for optical flow using pyramid, warping, and cost
  volume.
\newblock In {\em Proceedings of the IEEE conference on computer vision and
  pattern recognition}, pages 8934--8943, 2018.

\bibitem{waymo}
Pei Sun, Henrik Kretzschmar, Xerxes Dotiwalla, Aurelien Chouard, Vijaysai
  Patnaik, Paul Tsui, James Guo, Yin Zhou, Yuning Chai, Benjamin Caine, Vijay
  Vasudevan, Wei Han, Jiquan Ngiam, Hang Zhao, Aleksei Timofeev, Scott
  Ettinger, Maxim Krivokon, Amy Gao, Aditya Joshi, Yu Zhang, Jonathon Shlens,
  Zhifeng Chen, and Dragomir Anguelov.
\newblock Scalability in perception for autonomous driving: Waymo open dataset.
\newblock In {\em IEEE/CVF Conference on Computer Vision and Pattern
  Recognition (CVPR)}, June 2020.

\bibitem{raft}
Zachary Teed and Jia Deng.
\newblock Raft: Recurrent all-pairs field transforms for optical flow.
\newblock {\em arXiv preprint arXiv:2003.12039}, 2020.

\bibitem{attention}
Ashish Vaswani, Noam Shazeer, Niki Parmar, Jakob Uszkoreit, Llion Jones,
  Aidan~N Gomez, {\L}ukasz Kaiser, and Illia Polosukhin.
\newblock Attention is all you need.
\newblock In {\em Advances in neural information processing systems}, pages
  5998--6008, 2017.

\bibitem{mots}
Paul Voigtlaender, Michael Krause, Aljosa Osep, Jonathon Luiten, Berin
  Balachandar~Gnana Sekar, Andreas Geiger, and Bastian Leibe.
\newblock Mots: Multi-object tracking and segmentation.
\newblock In {\em Proceedings of the IEEE conference on computer vision and
  pattern recognition}, pages 7942--7951, 2019.

\bibitem{trackcolor}
Carl Vondrick, Abhinav Shrivastava, Alireza Fathi, Sergio Guadarrama, and Kevin
  Murphy.
\newblock Tracking emerges by colorizing videos.
\newblock In {\em Proceedings of the European conference on computer vision
  (ECCV)}, pages 391--408, 2018.

\bibitem{corrnet}
Heng Wang, Du Tran, Lorenzo Torresani, and Matt Feiszli.
\newblock Video modeling with correlation networks.
\newblock In {\em Proceedings of the IEEE/CVF Conference on Computer Vision and
  Pattern Recognition}, pages 352--361, 2020.

\bibitem{axial}
Huiyu Wang, Yukun Zhu, Bradley Green, Hartwig Adam, Alan Yuille, and
  Liang-Chieh Chen.
\newblock Axial-deeplab: Stand-alone axial-attention for panoptic segmentation.
\newblock {\em arXiv preprint arXiv:2003.07853}, 2020.

\bibitem{nonlocal}
Xiaolong Wang, Ross Girshick, Abhinav Gupta, and Kaiming He.
\newblock Non-local neural networks.
\newblock In {\em Proceedings of the IEEE conference on computer vision and
  pattern recognition}, pages 7794--7803, 2018.

\bibitem{jdmotgnn}
Yongxin Wang, Xinshuo Weng, and Kris Kitani.
\newblock Joint detection and multi-object tracking with graph neural networks.
\newblock {\em arXiv preprint arXiv:2006.13164}, 2020.

\bibitem{jde}
Zhongdao Wang, Liang Zheng, Yixuan Liu, and Shengjin Wang.
\newblock Towards real-time multi-object tracking.
\newblock In {\em European Conference on Computer Vision}, 2020.

\bibitem{deepsort}
Nicolai Wojke, Alex Bewley, and Dietrich Paulus.
\newblock Simple online and realtime tracking with a deep association metric.
\newblock In {\em 2017 IEEE international conference on image processing
  (ICIP)}, pages 3645--3649. IEEE, 2017.

\bibitem{cuhksysu}
Tong Xiao, Shuang Li, Bochao Wang, Liang Lin, and Xiaogang Wang.
\newblock Joint detection and identification feature learning for person
  search.
\newblock In {\em Proceedings of the IEEE Conference on Computer Vision and
  Pattern Recognition}, pages 3415--3424, 2017.

\bibitem{strmot}
Jiarui Xu, Yue Cao, Zheng Zhang, and Han Hu.
\newblock Spatial-temporal relation networks for multi-object tracking.
\newblock In {\em Proceedings of the IEEE International Conference on Computer
  Vision}, pages 3988--3998, 2019.

\bibitem{dnonlocal}
Minghao Yin, Zhuliang Yao, Yue Cao, Xiu Li, Zheng Zhang, Stephen Lin, and Han
  Hu.
\newblock Disentangled non-local neural networks.
\newblock {\em arXiv preprint arXiv:2006.06668}, 2020.

\bibitem{dilated}
Fisher Yu, Vladlen Koltun, and Thomas Funkhouser.
\newblock Dilated residual networks.
\newblock In {\em Proceedings of the IEEE conference on computer vision and
  pattern recognition}, pages 472--480, 2017.

\bibitem{poi}
Fengwei Yu, Wenbo Li, Quanquan Li, Yu Liu, Xiaohua Shi, and Junjie Yan.
\newblock Poi: Multiple object tracking with high performance detection and
  appearance feature.
\newblock In {\em European Conference on Computer Vision}, pages 36--42.
  Springer, 2016.

\bibitem{dla}
Fisher Yu, Dequan Wang, Evan Shelhamer, and Trevor Darrell.
\newblock Deep layer aggregation.
\newblock In {\em Proceedings of the IEEE conference on computer vision and
  pattern recognition}, pages 2403--2412, 2018.

\bibitem{fairmot}
Yifu Zhan, Chunyu Wang, Xinggang Wang, Wenjun Zeng, and Wenyu Liu.
\newblock A simple baseline for multi-object tracking.
\newblock {\em arXiv preprint arXiv:2004.01888}, 2020.

\bibitem{stability}
Hong Zhang and Naiyan Wang.
\newblock On the stability of video detection and tracking.
\newblock {\em arXiv preprint arXiv:1611.06467}, 2016.

\bibitem{flowfuse}
Jimuyang Zhang, Sanping Zhou, Xin Chang, Fangbin Wan, Jinjun Wang, Yang Wu, and
  Dong Huang.
\newblock Multiple object tracking by flowing and fusing.
\newblock {\em arXiv preprint arXiv:2001.11180}, 2020.

\bibitem{costflow}
Li Zhang, Yuan Li, and Ramakant Nevatia.
\newblock Global data association for multi-object tracking using network
  flows.
\newblock In {\em 2008 IEEE Conference on Computer Vision and Pattern
  Recognition}, pages 1--8. IEEE, 2008.

\bibitem{cityperson}
Shanshan Zhang, Rodrigo Benenson, and Bernt Schiele.
\newblock Citypersons: A diverse dataset for pedestrian detection.
\newblock In {\em Proceedings of the IEEE Conference on Computer Vision and
  Pattern Recognition}, pages 3213--3221, 2017.

\bibitem{prw}
Liang Zheng, Hengheng Zhang, Shaoyan Sun, Manmohan Chandraker, Yi Yang, and Qi
  Tian.
\newblock Person re-identification in the wild.
\newblock In {\em Proceedings of the IEEE Conference on Computer Vision and
  Pattern Recognition}, pages 1367--1376, 2017.

\bibitem{centertrack}
Xingyi Zhou, Vladlen Koltun, and Philipp Kr{\"a}henb{\"u}hl.
\newblock Tracking objects as points.
\newblock In {\em European Conference on Computer Vision}, 2020.

\bibitem{centernet}
Xingyi Zhou, Dequan Wang, and Philipp Kr{\"a}henb{\"u}hl.
\newblock Objects as points.
\newblock In {\em arXiv preprint arXiv:1904.07850}, 2019.

\bibitem{dcn}
Xizhou Zhu, Han Hu, Stephen Lin, and Jifeng Dai.
\newblock Deformable convnets v2: More deformable, better results.
\newblock In {\em Proceedings of the IEEE Conference on Computer Vision and
  Pattern Recognition}, pages 9308--9316, 2019.

\end{thebibliography}
}


\end{document}